\documentclass{egpubl}
\usepackage{eg2019}

\ConferencePaper        
 \electronicVersion 

\ifpdf \usepackage[pdftex]{graphicx} \pdfcompresslevel=9
\else \usepackage[dvips]{graphicx} \fi

\PrintedOrElectronic

\usepackage{t1enc,dfadobe}

\usepackage{egweblnk}
\usepackage{cite}

\usepackage[english]{babel}
\usepackage{booktabs} 
\usepackage{wrapfig}
\usepackage{subcaption}
\usepackage{stackengine}
\usepackage[normalem]{ulem}
\usepackage{algorithm}
\usepackage{algpseudocode}
\usepackage{tikz}
\usepackage{xcolor} 

\newcommand{\citet}[1]{\cite{#1}} 
\usepackage{amssymb} 
\usepackage{mathtools} 
\renewcommand{\paragraph}[1]{\emph{#1.}} 

\renewcommand{\vec}[1]{\mathbf{#1}}

\newcommand{\todo}[1]{}

\newcommand\hide[1]{}

\definecolor{bkcolor}{RGB}{210,10,210}
\definecolor{vcacolor}{RGB}{123,50,210}
\definecolor{nilscolor}{RGB}{10,110,240}
\definecolor{tedcolor}{RGB}{90,200,80}
\definecolor{barbaracolor}{RGB}{246,150,70}

\newcommand\ie{i.e.,~}
\newcommand\eg{e.g.,~}
\newcommand\Fig[1]{Figure~\ref{fig:#1}}
\newcommand\Sec[1]{Section~\ref{sec:#1}}
\newcommand\Eq[1]{Equation~(\ref{eq:#1})}

\newcommand\gradU{\nabla \vec{u}}

\newcommand\dataset{data set}
\newcommand\datasets{data sets}
\newcommand\twoD{2-D}
\newcommand\threeD{3-D}
\newcommand\maxSpeedup{700$\times$}
\newcommand\maxCompression{1300$\times$}
\newcommand\roms{reduced-order methods}

\newcommand\RoMs{Reduced-order Methods}
\newcommand\cnnName{Deep Fluids}

\newcommand{\Vdim}{V_{\textrm{dim}}}
\newcommand{\Gdim}{G_{\textrm{dim}}}
\newcommand{\dmax}{d_{\textrm{max}}}

\title[Deep Fluids: A Generative Network for Parameterized Fluid Simulations]%
      {Deep Fluids: A Generative Network for Parameterized Fluid Simulations}

\author[B. Kim et al.]
{\parbox{\textwidth}{\centering
Byungsoo Kim$^{1}$, Vinicius C. Azevedo$^{1}$, Nils Thuerey$^{2}$, Theodore Kim$^{3}$, Markus Gross$^{1}$ and Barbara Solenthaler$^{1}$}
\\
{\parbox{\textwidth}{
\centering
$^1$ETH Z{\"u}rich \quad $^2$Technical University of Munich \quad $^3$Pixar Animation Studios
}}}

\begin{document}


\teaser{
	\centering
	\includegraphics[trim=0cm 2cm 0cm 3cm, clip, width=1\textwidth]{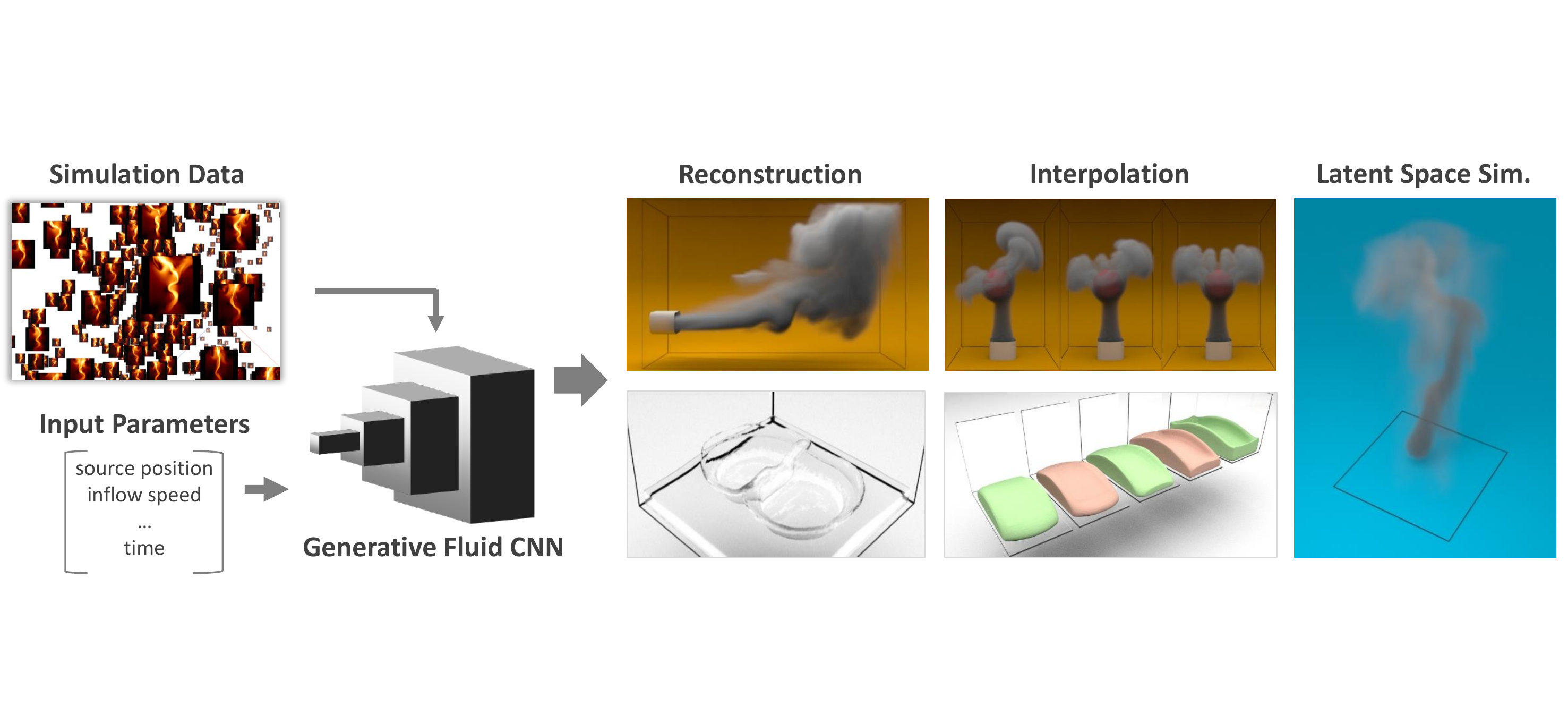}
	\caption{Our generative neural network synthesizes fluid velocities continuously in space and time, using a set of input simulations for training and a few parameters for generation. This enables fast reconstruction of velocities, continuous interpolation and latent space simulations.}
	\label{fig:teaser}
}

\maketitle
\begin{abstract}
This paper presents a novel generative model to synthesize fluid simulations from a set of reduced parameters. A convolutional neural network is trained on a collection of discrete, parameterizable fluid simulation velocity fields. Due to the capability of deep learning architectures to learn representative features of the data, our generative model is able to accurately approximate the training \dataset, while providing plausible interpolated in-betweens. The proposed generative model is optimized for fluids by a novel loss function that guarantees divergence-free velocity fields at all times. In addition, we demonstrate that we can handle complex parameterizations in reduced spaces, and advance simulations in time by integrating in the latent space with a second network. Our method models a wide variety of fluid behaviors, thus enabling applications such as fast construction of simulations, interpolation of fluids with different parameters, time re-sampling, latent space simulations, and compression of fluid simulation data. Reconstructed velocity fields are generated up to \maxSpeedup~faster than re-simulating the data with the underlying CPU solver, while achieving compression rates of up to \maxCompression.


\end{abstract}

\section{Introduction}
\label{sec:Introduction}

Machine learning techniques have become pervasive in recent years due to numerous algorithmic advances and the accessibility of computational power. Accordingly, they have been adopted for many applications in graphics, such as generating terrains ~\cite{guerin2017}, high-resolution faces synthesis~\cite{Karras2017} and cloud rendering~\cite{Kallweit2017}. In fluid simulation, machine learning techniques have been used to replace~\cite{Ladicky2015}, speed up~\cite{Tompson2016} or enhance existing solvers~\cite{Xie2018}.

Given the amount of available fluid simulation data, data-driven approaches have emerged as attractive solutions. Subspace solvers \cite{Treuille2006}, fluid re-simulators \cite{Kim2013} and basis compressors \cite{Jones2016} are examples of recent efforts in this direction. However, these methods usually represent fluids using linear basis functions, \eg constructed via Singular Value Decomposition (SVD), which are less efficient than their non-linear counterparts. In this sense, deep generative models implemented by convolutional neural networks (CNNs) show promise for representing data in reduced dimensions due to their capability to tailor non-linear functions to input data.
\begin{figure*}[t!]
    \centering
    \includegraphics[width=0.48\textwidth]{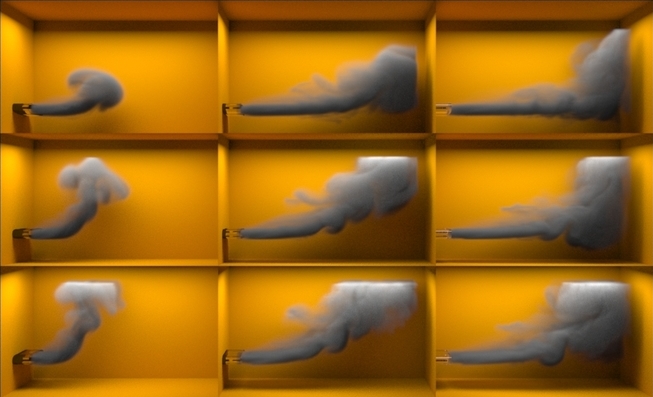}
    \hspace{4pt}
    \includegraphics[width=0.48\textwidth]{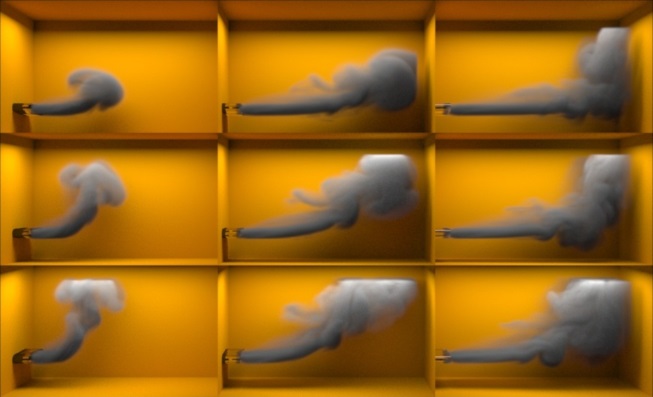}
    \caption{Ground truth (left) and the CNN-reconstructed results (right) for nine sample simulations with varying buoyancy (rows) and inflow velocity (columns). 
    Despite the varying dynamics of the ground truth simulations, our trained model closely reconstructs the reference data.}
    \label{fig:smokeGunArray}
    \vspace{-12pt}
\end{figure*}

In this paper, we propose the first generative neural network that fully constructs dynamic Eulerian fluid simulation velocities from a set of reduced parameters. Given a set of discrete, parameterizable simulation examples, our deep learning architecture generates velocity fields that are incompressible by construction. In contrast to previous subspace methods \cite{Kim2013}, our network achieves a wide variety of fluid behaviors, ranging from turbulent smoke to gooey liquids (Figure 1).

The \cnnName~CNN enhances the state of the art of \roms~(ROMs) in four ways: efficient evaluation time, a natural non-linear representation for interpolation, data compression capability and a novel approach for latent space simulations. Our CNN can generate a full velocity field in constant time, contrasting with previous approaches which are only efficient for sparse reconstructions \cite{Treuille2006}. Thanks to its \maxSpeedup~speed\-up compared to regular simulations, our approach is particularly suitable for animating physical phenomena in real-time applications such as games, VR and surgery simulators.

Our method is not only capable of accurately and efficiently recovering learned fluid states, but also generates plausible velocity fields for input parameters that have no direct correspondence in the training data. This is possible due to the inherent capability of deep learning architectures to learn representative features of the data. Having a smooth velocity field reconstruction when continuously exploring the parameter space enables various applications that are particularly useful for the prototyping of expensive fluid simulations: fast construction of simulations, interpolation of fluids with different parameters, and time re-sampling.
To handle applications with extended parameterizations such as the moving smoke scene shown in Section \ref{sec:3dsmoke}, we couple an encoder architecture with a latent space integration network. This allows us to advance a simulation in time by generating a sequence of suitable latent codes.
Additionally, the proposed architecture works as a powerful compression algorithm for velocity fields with compression rates that outperform previous work~\cite{Jones2016} by two orders of magnitude.
%

To summarize, the technical contributions of our work include:
\begin{itemize}
  \item The first generative deep learning architecture that fully synthesizes plausible and fully divergence-free \twoD~and \threeD~fluid simulation velocities from a set of reduced parameters.
  \item A generative model for fluids that accurately encodes parameterizable velocity fields. Compression rates of \maxCompression~are achieved, as well as \maxSpeedup~performance speed-ups compared to using the underlying CPU solver for re-simulation.
  \item An approach to encode simulation classes into a latent space representation through an autoencoder architecture. In combination with a latent space integration network to advance time, our approach allows flexible interactions with flow simulations.
  \item A detailed analysis of the proposed method, both when reconstructing samples that have a direct correspondence with the training data and intermediate points in the parameter space.
\end{itemize}

\section{Related Work}
\label{sec:RelatedWork}

\paragraph{\RoMs} Subspace solvers aim to accelerate simulations by discovering simplified representations. In engineering, these techniques go back decades \cite{Lumley:1967}, but were introduced to computer graphics by \citet{Treuille2006} and \citet{Gupta:2007}. Since then, improvements have been made to make them modular \cite{Wicke2009}, consistent with widely-used integrators \cite{Kim2013}, more energy-preserving \cite{Liu:2015}\ and memory-efficient \cite{Jones2016}. A related "Laplacian Eigenfunctions'' approach \cite{deWitt2012} has also been introduced and refined \cite{Gerszewski:2015}, removing the need for snapshot training data when computing the linear subspace. The basis functions used by these methods are all linear however, and various methods are then used to coerce the state of the system onto some non-linear manifold. Exploring the use of non-linear functions, as we do here, is a natural evolution.
One well-known limitation of \roms~is their inability to simulate liquids because the non-linearity of the liquid interface causes the subspace dimensionality to explode. For example, in solid-fluid coupling, usually the fluid is computed directly while only the solid uses the reduced model \cite{Lu:2016:TCF}. Graph-based methods for precomputing liquid motions \cite{Stanton:2014} have had some success, but only under severe constraints, e.g.~the user viewpoint must be fixed. In contrast, we show that the non-linearity of a CNN-based approach allows it to be applied to liquids as well.

\paragraph{Machine Learning \& Fluids}
Combining fluid solvers with machine learning techniques was first demonstrated by \citet{Ladicky2015}. By employing Regression Forests to approximate the Navier-Stokes equations on a Lagrangian system, particle positions and velocities were predicted with respect to input parameters for a next time step. Regression Forests are highly efficient, but require handcrafted features that lack the generality and abstraction power of CNNs.
An LSTM-based method for predicting changes of the pressure field for multiple subsequent time steps has been presented by \citet{Wiewel2018}, resulting in significant speed-ups of the pressure solver. For a single time step, a CNN-based pressure projection was likewise proposed \cite{Tompson2016, Yang2016}. In contrast to our work, these models only replace the pressure projection stage of the solver, and hence are specifically designed to accelerate the enforcement of divergence-freeness.
To visually enhance low resolution simulations, \citet{Chu2017} synthesized smoke details by looking up pre-computed patches using CNN-based descriptors, while \citet{Xie2018} proposed a GAN for super resolution smoke flows.
Other works enhance FLIP simulations with a learned splash model  \cite{Um2017}, while the deformation learning proposed by \cite{Prantl2017}
shares our goal of capturing sets of fluid simulations.
However, it only partially employs CNNs and focuses on signed distance functions, while our work targets the velocity spaces of a broad class of fluid simulations.
Lattice-Boltzmann steady-state flow solutions are recovered by CNN surrogates using signed distance functions as input boundary conditions in \cite{Guo2016}. \citet{Farimani2017} use Generative Adversarial Networks (GANs) \cite{GoodFellow2014} to train steady state heat conduction and steady incompressible flow solvers. Their method is only demonstrated in 2-D and the interpolation capabilities of their architecture are not explored. For both methods, the simulation input is a set of parameterized initial conditions defined over a spatial grid, and the output is a single steady state solution. Recently, \citet{Umetani2018} developed a data-driven technique to predict fluid flows around bodies for interactive shape design, while Ma et al. \shortcite{maDrl2018} have demonstrated deep learning based fluid interactions with rigid bodies.

\paragraph{Machine Learning \& Physics} In the physics community, neural networks and deep learning architectures for approximating, enhancing and modeling solutions to complex physics problems are gaining attention. A few recent examples are \citet{Carleo2016} using reinforcement learning to reduce the complexity of a quantum many-body problem, \citet{Ling2016} employing deep neural networks to synthesize Reynolds average turbulence anisotropy tensors from high-fidelity simulation data, 
and 
\citet{Paganini2017} modeling calorimeter interactions with electromagnetic showers using GANs.
GANs have also been employed to generate \cite{Ravanbakhsh2016} and deconvolve \cite{Schawinski2017} galaxy images, and reconstruct three-dimensional porous media \cite{Mosser2017}. As we focus on generative networks for known parameterizations, we will not employ learned, adversarial losses. Rather, we will demonstrate that a high quality representation can be learned by constructing a suitable direct loss function.

\section{A Generative Model For Fluids}
\label{sec:Method}
Fluids are traditionally simulated by solving the inviscid momentum $D\vec{u}/Dt = -\nabla p + \vec{g}$ and mass conservation $\nabla \cdot \vec{u} = 0$ equations, where $\vec{u}$ and $p$ are the fluid velocity and pressure, $D\vec{u}/Dt$ is the material derivative and $\vec{g}$ represents external forces. The viscosity $-\mu \nabla^2 \vec{u}$ can be included, but simulations for visual effects usually rely on numerical dissipation instead. For a given set of simulated fluid examples, our goal is to train a CNN that approximates the original velocity field data set.
By minimizing loss functions with subsequent convolutions applied to its input, CNNs organize the data manifold into shift-invariant feature maps.

Numerical fluid solvers work by advancing a set of fully specified initial conditions. By focusing on scenes that are initially parameterizable by a handful of variables, such as the position of a smoke source, we are able to generate samples for a chosen class of simulations. Thus, the inputs for our method are parameterizable \datasets, and we demonstrate that accurate generative networks can be trained in a supervised way.

\subsection{Loss Function for Velocity Reconstruction}
\begin{figure}[t!]
    \centering
    \includegraphics[width=0.075\textwidth]{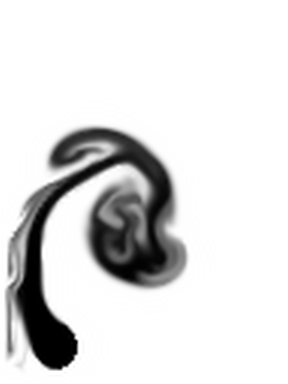}
    \includegraphics[width=0.075\textwidth]{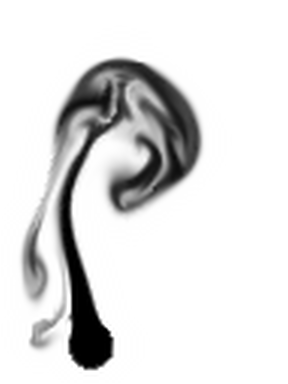}
    \includegraphics[width=0.075\textwidth]{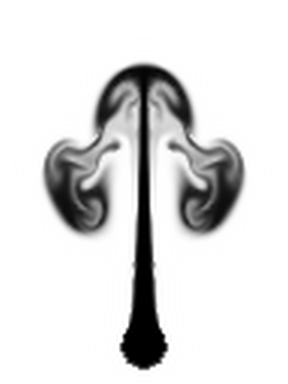}
    \includegraphics[width=0.075\textwidth]{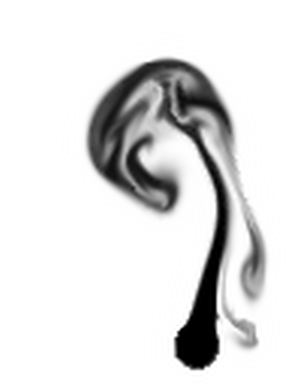}
    \includegraphics[width=0.075\textwidth]{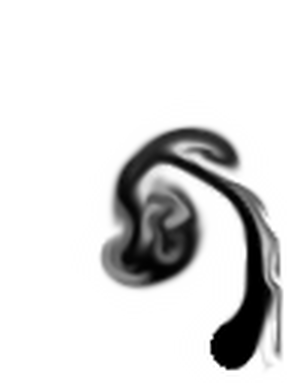}
    \\
    \includegraphics[width=0.075\textwidth]{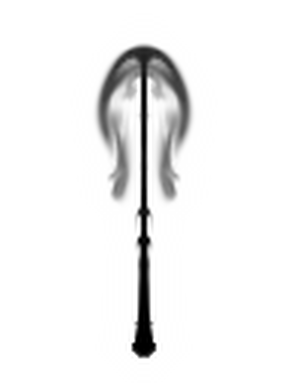}
    \includegraphics[width=0.075\textwidth]{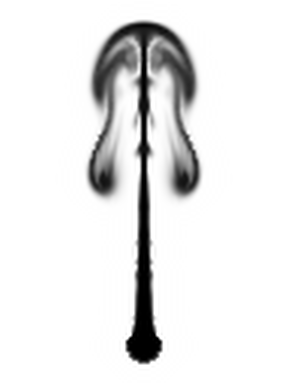}
    \includegraphics[width=0.075\textwidth]{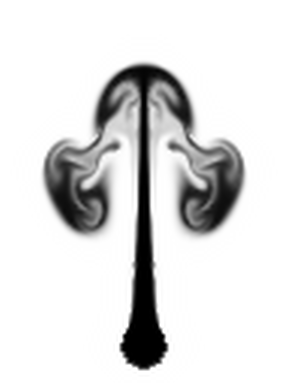}
    \includegraphics[width=0.075\textwidth]{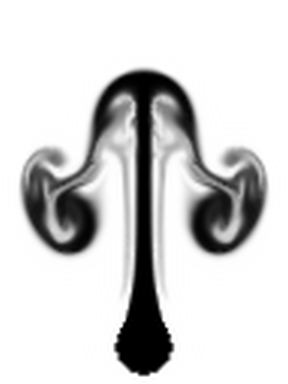}
    \includegraphics[width=0.075\textwidth]{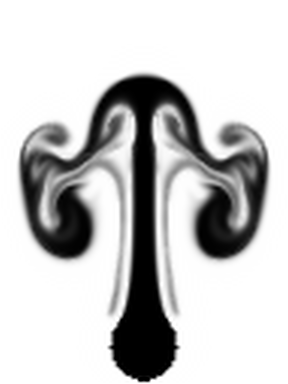}
    \caption{
    Different snapshots showing the advected densities for varying smoke source parameters. The top and bottom rows show the variation of the initial position source and width, respectively.
    %
     }
    \label{fig:smoke2DSetup}
    \vspace{-12pt}
\end{figure}

The network's input is characterized by a pair $[\vec{u}_c, \vec{c}]$, where $\vec{u}_c \in \mathbb{R}^{H \times W \times D \times \Vdim}$ is a single velocity vector field frame in $\Vdim$ dimensions (i.e. $\Vdim=2$ for \twoD~and $\Vdim=3$ for \threeD) with height $H$, width $W$ and depth $D$ (1 for~\twoD), generated using the solver's parameters $\vec{c} = [c_1, c_2, ..., c_n] \in \mathbb{R}^n$. For the \twoD~example in \Fig{smoke2DSetup}, $\vec{c}$ is the combination of $x$-position and width of the smoke source, and the current time of the frame. Due to the inherent non-linear nature of the Navier-Stokes equations, these three parameters (i.e.~position, width, and time) yield a vastly different set of velocity outputs.

For fitting fluid samples, our network uses velocity-parameter pairs and updates its internal weights by minimizing a loss function. This process is repeated by random batches until the network minimizes a loss function over all the training data. While previous works have proposed loss functions for natural images, e.g., $L_p$ norms, MS-SSIM~\cite{ZhaoGFK15}, and perceptual losses~\cite{Johnson2016,Ledig2016}, accurate reconstructions of velocity fields have not been investigated. For fluid dynamics it is especially important to ensure conservation of mass,
i.e., to ensure divergence-free motion for incompressible flows. We therefore propose a novel {\em stream function} based loss function defined as
\begin{equation}
L_G (\vec{c}) = ||\vec{u}_\vec{c} - \nabla \times G(\vec{c})||_1.
\label{eq:lossdivfree}
\end{equation}
$G(\vec{c}): \mathbb{R}^{n} \mapsto \mathbb{R}^{H \times W \times D \times \Gdim}$ is the network output and $\vec{u}_\vec{c}$ is a simulation sample from the training data. The curl of the model output is the reconstruction target, and it is \emph{guaranteed} to be divergence-free by construction, as $\nabla \cdot (\nabla \times G(\vec{c})) = 0$. Thus, $G(\vec{c})$  implicitly learns to approximate a stream function $\Psi_\vec{c}$ 
(i.e. $\Gdim=1$ for \twoD~and $\Gdim=3$ for \threeD) that corresponds to a velocity sample $\vec{u}_\vec{c}$. 

While this formulation is highly suitable for incompressible flows, regions with partially divergent motion, such as extrapolated velocities around the free surface of a liquid, are better approximated with a direct velocity inference. For such cases, we remove the curl term from \Eq{lossdivfree}, and instead use 
\begin{equation}
L_G (\vec{c}) = ||\vec{u}_\vec{c} - G(\vec{c})||_1
\end{equation}
where the output of $G$ represents a velocity field with $G(\vec{c}): \mathbb{R}^{n} \mapsto \mathbb{R}^{H \times W \times D \times \Vdim}$.

\setlength{\columnsep}{0pt}%
\begin{wrapfigure}[10]{hR}{0.35\linewidth} 
\vspace{-4mm}
\begin{center}
\includegraphics[width=0.65\linewidth]{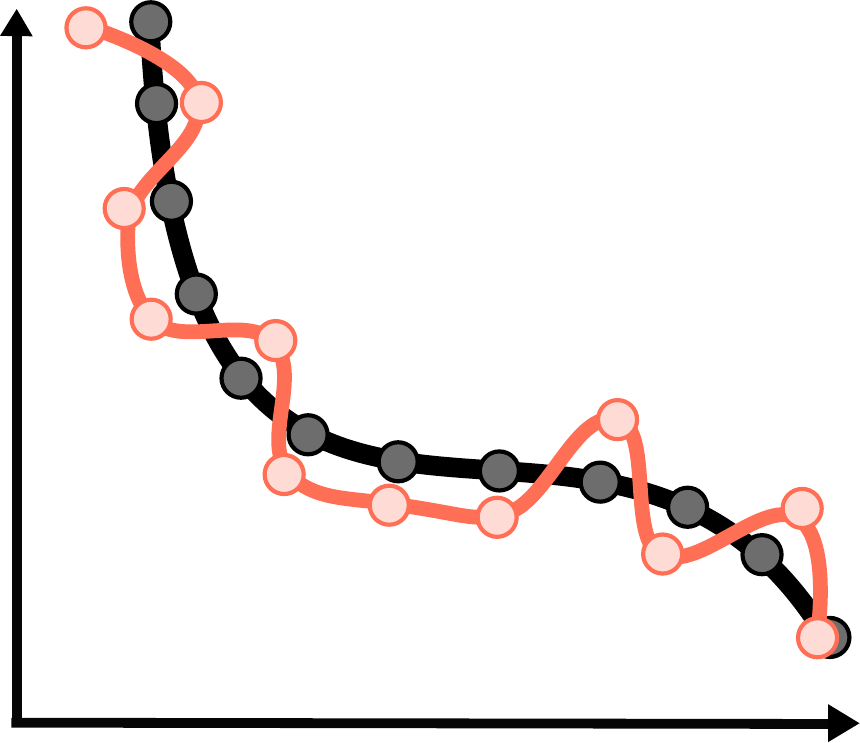}
 \smallskip\par
\includegraphics[width=0.65\linewidth]{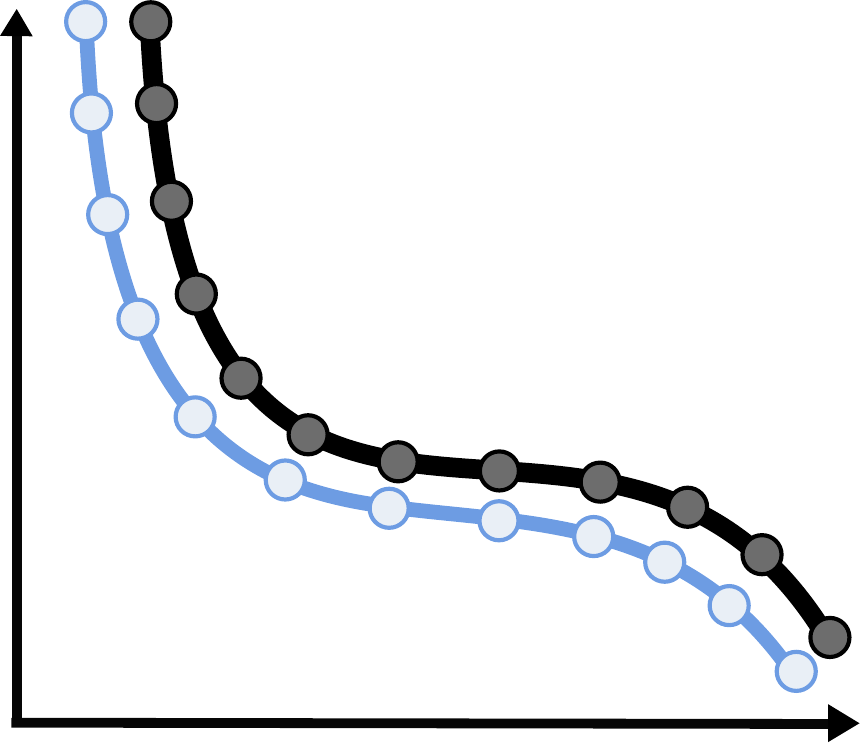}
\end{center}
\vspace{-4mm}
\label{fig:insetGradComparision}
\end{wrapfigure}

In both cases, simply minimizing the $L_1$ distance of a high-order function approximation to its original counterpart does not guarantee that their derivatives will match. Consider the example shown in the inset image: given a function (black line, gray circles), two approximations (red line, top image; blue line, bottom image) of it with the same average $L_1$ distances are shown. In the upper image derivatives do not match, producing a jaggy reconstructed behavior; in the bottom image both values and derivatives of the $L_1$ distance are minimized, resulting in matching derivatives. With a sufficiently smooth data set, high-frequency features of the CNN are in the null space of the $L_1$ distance minimization and noise may occur. We discuss this further in the supplemental material.

Thus, we augment our loss function to also minimize the difference of the velocity field gradients. The velocity gradient $\nabla: \mathbb{R}^{H \times W \times D \times \Vdim} \mapsto \mathbb{R}^{H \times W \times D \times (\Vdim)^2 }$ is a second-order tensor that encodes vorticity, shearing and divergence information. Similar techniques, as image gradient difference loss \cite{Mathieu2015}, have been employed for improving frame prediction on video sequences. However, to our knowledge, this is the first architecture to employ gradient information to improve velocity field data. Our resulting loss function is defined as
\begin{equation}
L_G (\vec{c}) = \lambda_{\vec{u}}||\vec{u}_\vec{c} - \hat{\vec{u}}_\vec{c}||_1 + \lambda_{\gradU}||\nabla \vec{u}_\vec{c} - \nabla \hat{\vec{u}}_\vec{c}||_1,
\label{eq:lossDiscreteGenerator}
\end{equation}
where $\hat{\vec{u}}_\vec{c} = \nabla \times G(\vec{c})$ for incompressible flows and $\hat{\vec{u}}_\vec{c} =G(\vec{c})$ for compressible flows, and $\lambda_{\vec{u}}$ and $\lambda_{\gradU}$ are weights used to emphasize the reconstruction of either the velocities or their derivatives. In practice, we used $\lambda_{\vec{u}} = \lambda_{\gradU} = 1$ for normalized data (see~\Sec{Training}). The curl of the velocity and its gradients are computed internally by our architecture and do not need to be explicitly included in the data set.

\subsection{Implementation}
\begin{figure}[t!]
    \centering
    \includegraphics[width=0.48\textwidth]{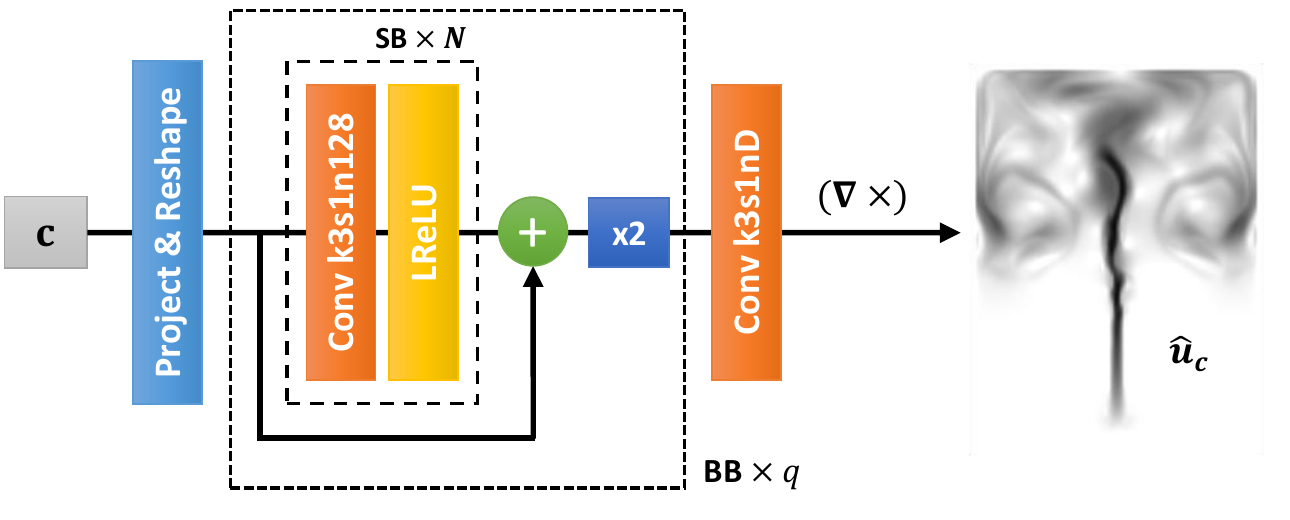}
    \caption{Architecture of the proposed generative model, subdivided into small ($SB$) and big blocks ($BB$). Small blocks are composed of flat convolutions followed by a LReLU activation function. Big blocks are composed of sets of small blocks, an additive skip-connection and an upsampling operation. The output of the last layer has $D$ channels ($\Gdim$ for incompressible velocity fields, $\Vdim$ otherwise) corresponding to the simulation dimension.}
    \label{fig:networkArchitecture}
    \vspace{-12pt}
\end{figure}

For the implementation of our generative model we adopted and modified the network architecture from~\cite{Berthelot2017}. As illustrated in~\Fig{networkArchitecture}, our generator starts by projecting the initial $\vec{c}$ parameters into an $m$-dimensional vector of weights $\vec{m}$ via fully connected layers. The dimension of $\vec{m}$ depends on the network output $\vec{d} = [H, W, D, \Vdim]$ and on a custom defined parameter $q$. With $\dmax = \max(H, W, D)$, $q$ is calculated by $q = \log_2(\dmax) - 3$, meaning that the minimum supported number of cells in one dimension is 8. Additionally, we constrain all our grid dimensions to be divisible by $2^q$. Since we use a fixed number of feature maps per layer, the number of dimensions of $\vec{m}$ is $m = \frac{H}{2^q}\times \frac{W}{2^q} \times \frac{D}{2^q} \times 128$ and those will be expanded to match the original output resolution.

The $m$-dimensional vector $\vec{m}$ is then reshaped to a $[\frac{H}{2^q}, \frac{W}{2^q}, \frac{D}{2^q}, 128]$ tensor. As shown in \Fig{networkArchitecture}, the generator component is subdivided into small ($SB$) and big blocks ($BB$). For small blocks, we perform $N$ (most of our examples used $N = 4~\text{or}~5$) flat convolutions  followed by Leaky Rectified Linear Unit (LReLU) activation functions~\cite{maas2013lrelu}.
We substituted the Exponential Liner Unit (ELU) activation function in the original method from~\cite{Berthelot2017} by the LReLU as it yielded sharper outputs when minimizing the $L_1$ loss function. Additionally, we employ residual skip connections~\cite{He2015}, which are an element-wise sum of feature maps of input and output from each $SB$. While the concatenative skip connections employed by \citet{Berthelot2017} are performed between the first hidden states and the consecutive maps with doubling of the depth size to 256, ours are applied to all levels of $SB$s with a fixed size of 128 feature maps.  After the following upsample operation, the dimension of the output from a $BB$ after $i$ passes is $[\frac{H}{2^{q-i}}, \frac{W}{2^{q-i}}, \frac{D}{2^{q-i}}, 128]$. Our experiments showed that performing these reductions to the feature map sizes with the residual concatenation improved the network training time without degradation of the final result.

\section{Extended Parameterizations}

Scenes with a large number of parameters can be challenging to parameterize. For instance, the dynamically moving smoke source example (\Fig{torchresim}) can be parameterized by the history of control inputs, \ie $[\vec{p}_0, \vec{p}_1, ..., \vec{p}_t] \rightarrow \vec{u}_t$, where $\vec{p}_t$ and $\vec{u}_t$ represent the smoke source position and the reconstructed velocity field at time $t$, respectively. In this case, however, the number of parameters grows linearly with the number of frames tracking user inputs.
As a consequence, the parameter space would be infeasibly large for data-driven approaches, and would be extremely costly to cover with samples for generating training data.

To extend our approach to these challenging scenarios, we add an encoder architecture $G^{\dagger}(\vec{u}) : \mathbb{R}^{H \times W \times D \times \Vdim} \mapsto \mathbb{R}^{n}$ to our generator of Section \ref{sec:Method}, 
and combine it with a second smaller network
for time integration (Section \ref{sec:integrationnetwork}), as illustrated in \Fig{architecture_ae}.
%
In contrast to our generative network, the encoder architecture maps velocity field frames into a parameterization $\vec{c} = [\vec{z}, \vec{p}] \in \mathbb{R}^n$, in which $\vec{z} \in \mathbb{R}^{n-k}$ is a reduced latent space that models arbitrary features of the flow in an unsupervised way and $\vec{p} \in \mathbb{R}^{k}$ is a supervised parameterization to control specific attributes~\cite{Kulkarni2015}. Note that this separation makes the latent space sparser while training, which in turn improves the quality of the reconstruction.
For the moving smoke source example in~\Sec{3dsmoke}, $n=16$ and $\vec{p}$ encodes $x,z$ positions used to control the position of the smoke source.

The combined encoder and generative networks are similar to Deep Convolutional autoencoders~\cite{Vincent2010}, where the generative network $G(\mathbf{c})$ acts as a decoder. The encoding architecture is symmetric to our generative model, except that we do not employ the inverse of the curl operator and the last convolutional layer. We train both generative and encoding networks with a combined loss similar to \Eq{lossDiscreteGenerator}, as
\begin{equation}
L_{AE} (\vec{u}) = \lambda_{\vec{u}}||\vec{u}_\vec{c} - \hat{\vec{u}}_\vec{c}||_1 + \lambda_{\gradU}||\nabla \vec{u}_\vec{c} - \nabla \hat{\vec{u}}_\vec{c}||_1 + \lambda_{\vec{p}}||\vec{p} - \hat{\vec{p}}||^2_2,
\label{eq:lossae}
\end{equation}
where $\hat{\vec{p}}$ is the part of the latent space vector constrained to represent control parameters $\vec{p}$, and $\lambda_{\vec{p}}$ is a weight to emphasize the learning of supervised parameters. Note that the $L_2$ distance is applied to control parameters unlike vector field outputs, as it is a standard cost function in linear regression. As before, we used $\lambda_{\vec{u}} = \lambda_{\gradU} = \lambda_{\vec{p}} = 1$ for all our normalized examples~(\Sec{Training}). With this approach we can handle complex parameterizations, since the velocity field states are represented by the remaining latent space dimensions in  $\vec{z}$. This allows us to use latent spaces which do not explicitly encode the time dimension as a parameter. Instead, we can use a second latent space integration network that generates a suitable sequence of latent codes.

\subsection{Latent Space Integration Network}
\label{sec:integrationnetwork}
\begin{figure}[h]
    \centering
    \includegraphics[width=0.48\textwidth]{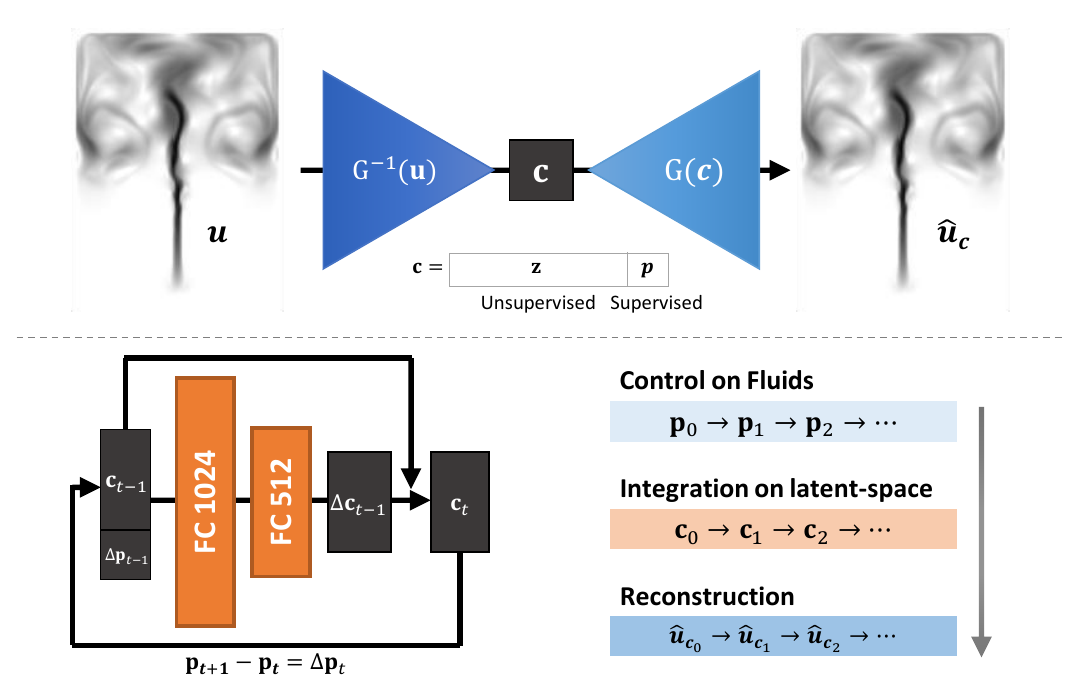}
    \caption{Autoencoder (top) and latent space integration network (bottom). The autoencoder compresses a velocity field $\vec{u}$ into a latent space representation $\vec{c}$, which includes a supervised and unsupervised part ($\vec{p}$ and $\vec{z}$). The latent space integration network finds mappings from subsequent latent code representations $\vec{c}_t$ and $\vec{c}_{t+1}$.}
    \label{fig:architecture_ae}
    \vspace{-12pt}
\end{figure}

The latent space only learns a diffuse representation of time by the velocity field states $\vec{z}$. Thus we propose a latent space integration network for advancing time from reduced representations.
The network $T(\vec{x}_t) : \mathbb{R}^{n+k} \mapsto \mathbb{R}^{n-k} $ takes an input vector $\vec{x}_t = [\vec{c}_t;\Delta\vec{p}_t] \in \mathbb{R}^{n + k}$ which is a concatenation of a latent code $\vec{c}_t$ at current time $t$ and a control vector difference between user input parameters $\Delta\vec{p}_t=\vec{p}_{t+1}-\vec{p}_t \in \mathbb{R}^{k}$. The parameter $\Delta\vec{p}_t$ has the same dimensionality $k$ as the supervised part of our latent space, and serves as a transition guidance from latent code $\vec{c}_t$ to $\vec{c}_{t+1}$. The output of $T(\vec{x}_t)$ is the residual $\Delta\vec{z}_{t}$ between two consecutive states. Thus, a new latent code is computed with $\vec{z}_{t+1} =\vec{z}_t + T(\vec{x}_t)$ as seen in~\Fig{architecture_ae}.

For improved accuracy we let $T$ look ahead in time, by training the network on a window of $w$ sequential latent codes with an $L_2$ loss function:
\begin{equation}
L_{T} (\vec{x}_t,...,\vec{x}_{t+w-1}) = \frac{1}{w} \sum^{t+w-1}_{i=t} ||\Delta\vec{z}_{i} - T_i||^2_2,
\label{eq:losstn}
\end{equation}
where $T_i$ is recursively computed from $t$ to $i$. Our window loss~\Eq{losstn} is designed to minimize not only errors on the next single step integration but also 
errors accumulated in repeated latent space updates. We found that $w=30$ yields good results, and a discussion of the effects of different values of $w$ is provided in the supplemental material.

We realize $T$ as a multilayer perceptron (MLP) network. The rationale behind choosing MLP instead of LSTM is that $T$ is designed to be a navigator on the manifold of the latent space, and we consider these integrations as controlled individual steps rather than physically induced ones. The network consists of three fully connected layers coupled with ELU activation functions. We employ batch normalization and dropout layers with probability of 0.1 to avoid overfitting.
Once the networks $G, G^{\dagger}$ and $T$ are trained, we use Algorithm~\ref{alg:algorithm} to reconstruct the velocity field for a new simulation. The algorithm starts from an initial reduced space that can be computed from an initial incompressible velocity field. The main loop consists of concatenating the reduced space and the position update into $\vec{x}_t$; then the latent space integration network computes $\Delta\vec{z}_t$, which is used to update $\vec{c}_t$ to $\vec{c}_{t+1}$. Finally, the generative network $G$ reconstructs the velocity field $\vec{u}_{t+1}$ by evaluating $\vec{c}_{t+1}$.

\begin{algorithm}
\caption{Simulation with the Latent Space Integration Network}
\begin{algorithmic}
\State{{$\vec{c_{0}} \gets G^{\dagger}(\vec{u_{0}})$}}
\While{simulating from $t$ to $t+1$}
  \State{$\vec{x}_t \gets [\vec{c}_t; \Delta\vec{p}_t] \quad$ // $\vec{c}_t$ from previous step, $\vec{p}$ is given} 
  \State{$\vec{z}_{t+1} \gets \vec{z}_t + T(\vec{x}_t)\quad$ // latent code inference }
  \State{$\vec{c}_{t+1} \gets [\vec{z}_{t+1}; \vec{p}_{t+1}]$}
  \State{$\vec{u}_{t+1} \gets G(\vec{c}_{t+1}) \quad$ // velocity field reconstruction}
\EndWhile
\end{algorithmic}
\label{alg:algorithm}
\end{algorithm}
\vspace{-0.2cm}

\section{Results}
In the following we demonstrate that our \cnnName\;CNN can reliably recover and synthesize dynamic flow fields for both smoke and liquids.
We refer the reader to the supplemental video for the corresponding animations.
For each scene, we reconstruct velocity fields computed by the generative network and advect densities for smoke simulations or surfaces for liquids. Vorticity confinement or turbulence synthesis were not applied after the network's reconstruction, but such methods could be added as a post-processing step. We trained our networks using the Adam optimizer~\cite{kingma2014adam} for 300,000 iterations with varying batch sizes to maximize GPU memory usage (8 for \twoD~and 1 for \threeD). For the time network $T$, we use 30,000 iterations. The learning rate of all networks is scheduled by a cosine annealing decay~\cite{Loshchilov2016}, where we use the learning range from \citet{Smith2015}. Scene settings, computation times and memory consumptions are summarized in Table~\ref{tab:stat}. Fluid scenes were computed with Mantaflow~\cite{mantaflow} using an Intel i7-6700K CPU at 4.00 GHz with 32GB memory, and CNN timings were evaluated on a 8GB NVIDIA GeForce GTX 1080 GPU. Networks are trained on a 12GB NVIDIA Titan X GPU. 

\subsection{\twoD~Smoke Plume}
\label{sec:smokePlume2D}
A sequence of examples which portray varying, rising smoke plumes in a rectangular container is shown in \Fig{smoke2DSetup}, where advected densities for different initial source positions (top) and widths (bottom) are shown. Since visualizing the advected smoke may result in blurred flow structures, we display vorticities instead, facilitating the understanding of how our CNN is able to reconstruct and interpolate between samples present in the \dataset. Additionally, we use the \emph{hat} notation to better differentiate parameters that do not have a direct correspondence with the ground truth data (\eg $\hat{\mathbf{p}}_x$ for an interpolated position on the x-axis). Our training set for this example consists of the combination of $5$ samples with varying source widths $w$ and $21$ samples with varying $x$ positions $\mathbf{p}_x$. Each simulation is computed for $200$ frames, using a grid resolution of $96\times128$ and a domain size of $(1, 1.\overline{33})$. The network is trained with a total of $21,000$ unique velocity field samples.

\begin{figure}[t!]
    \centering
    \stackunder{\includegraphics[trim = 0px 0px 0px 0px, clip, width=0.11\textwidth]{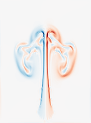}}{\scriptsize{G.t. $p_x = 0.5$}}
    \stackunder{\includegraphics[trim = 0px 0px 0px 0px, clip, width=0.11\textwidth]{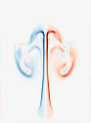}}{\scriptsize{CNN $p_x = 0.5$}}
    \stackunder{\includegraphics[trim = 0px 0px 0px 0px, clip, width=0.11\textwidth]{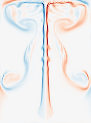}}{\scriptsize{G.t. $p_x = 0.5$}}
    \stackunder{\includegraphics[trim = 0px 0px 0px 0px, clip, width=0.11\textwidth]{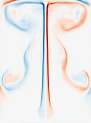}}{\scriptsize{CNN $p_x = 0.5$}}
    \caption{Vorticity plot of a \twoD~smoke simulation with direct correspondences to the training data set for two different times. The RdBu colormap is used to show both the magnitude and the rotation direction of the vorticity (red: clockwise). Our CNN is able to closely approximate ground truth samples (G.t.).}
    \label{fig:smokeVorticityComparison}
    \vspace{-12pt}
\end{figure}

\paragraph{Reconstruction with Direct Correspondences to the Data Set}
To analyze the reconstruction power of our approach, we compare generated velocities for parameters which have a direct correspondence to the original \dataset, i.e. the ground truth samples. \Fig{smokeVorticityComparison} shows vorticity plots comparing the ground truth (G.t.) and our CNN output for two different frames. The CNN shows a high reconstruction quality, where coarse structures are almost identically reproduced, and fine structures are closely approximated.

\paragraph{Sampling at Interpolated Parameters}
\begin{figure}[th!]
    \centering
    \includegraphics[width=0.10\textwidth]{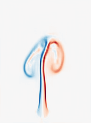}
    \includegraphics[width=0.10\textwidth]{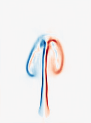}
    \includegraphics[width=0.10\textwidth]{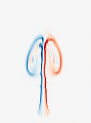}
    \includegraphics[width=0.10\textwidth]{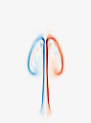}
    \vspace{1pt}
    \\
    \includegraphics[trim={0 5 0 0},clip,width=0.10\textwidth]{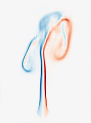}
    \includegraphics[trim={0 5 0 0},clip,width=0.10\textwidth]{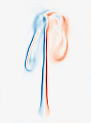}
    \includegraphics[trim={0 5 0 0},clip,width=0.10\textwidth]{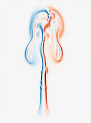}
    \includegraphics[trim={0 5 0 0},clip,width=0.10\textwidth]{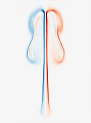}
    \vspace{1pt}
    \\
    \hspace{-4pt}
    \stackunder {\includegraphics[trim={0 5 0 0},clip,width=0.10\textwidth]{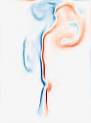}}{\scriptsize{CNN $p_x = 0.46$}}
    \stackunder {\includegraphics[trim={0 5 0 0},clip,width=0.10\textwidth]{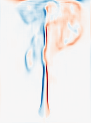}}{\scriptsize{CNN $\hat{p}_x = 0.48$}}
    \stackunder {\includegraphics[trim={0 5 0 0},clip,width=0.10\textwidth]{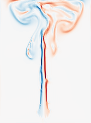}}{\scriptsize{G.t. ${p}_x = 0.48$}}
    \stackunder {\includegraphics[trim={0 5 0 0},clip,width=0.10\textwidth]{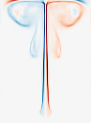}}{\scriptsize{CNN $p_x = 0.5$}}
    \caption{
    Vorticity plot of a \twoD~smoke simulation showing CNN reconstructions at ground truth correlated positions $\mathbf{p}_x = 0.46$ and $\mathbf{p}_x = 0.5$, the interpolated result at $\mathbf{\hat{p}}_x = 0.48$, and ground truth (G.t.) at $\mathbf{\hat{p}}_x = 0.48$ which is not part of the training \dataset.    
    }
    \label{fig:smokeVorticityInterpolation}
    \vspace{-15pt}
\end{figure}
We show the interpolation capability of our approach in \Fig{smokeVorticityInterpolation}. Left and right columns show the CNN reconstructions at ground truth correlated positions $\mathbf{p}_x = 0.46$ and $\mathbf{p}_x = 0.5$, while the second column shows a vorticity plot interpolated at $\mathbf{\hat{p}}_x = 0.48$. The third column shows the simulated ground truth for the same position.
For positions not present in the original data, our CNN synthesizes plausible new motions that are close to ground truth simulations.

\subsection{\threeD~Smoke Examples}
\label{sec:3dsmoke}
\begin{figure}[h]
    \centering 
    \includegraphics[width=0.11\textwidth]{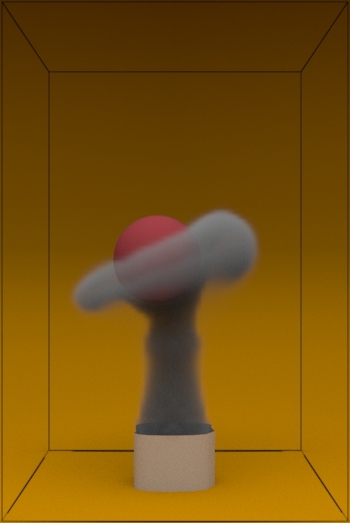}
    \includegraphics[width=0.11\textwidth]{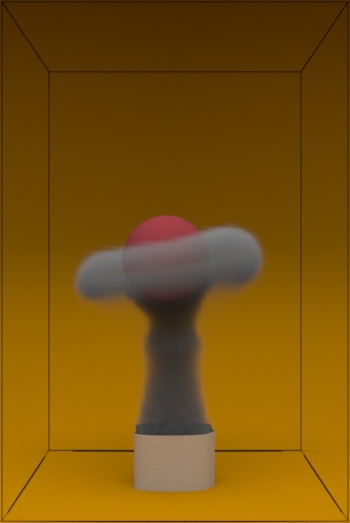}
    \includegraphics[width=0.11\textwidth]{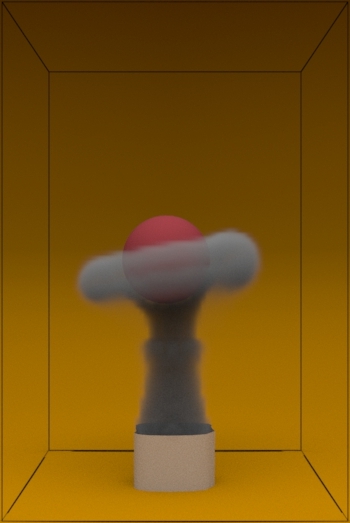}
    \includegraphics[width=0.11\textwidth]{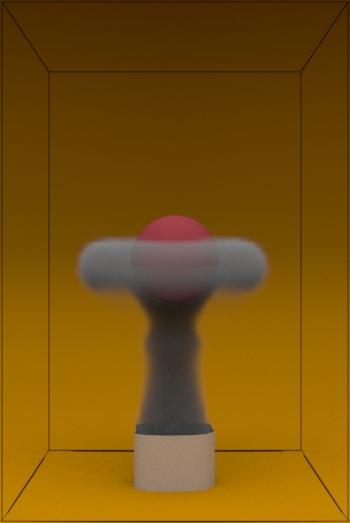}
    \vspace{1pt}
    \\
    \includegraphics[width=0.11\textwidth]{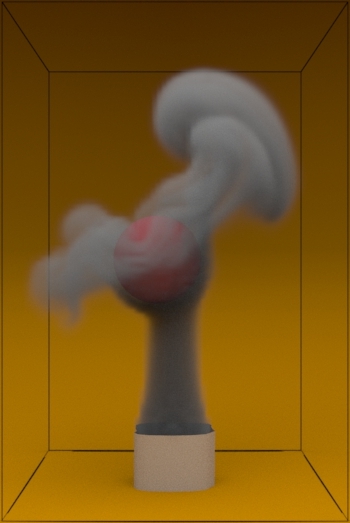}
    \includegraphics[width=0.11\textwidth]{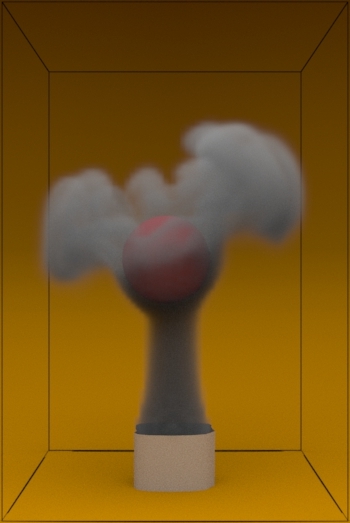}
    \includegraphics[width=0.11\textwidth]{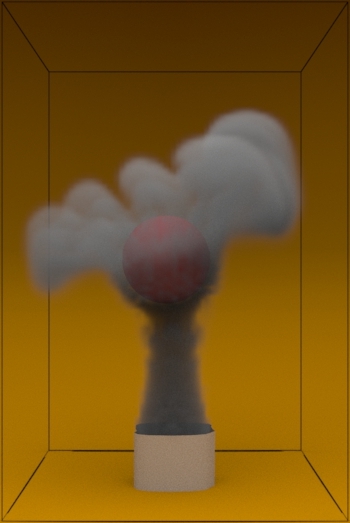}
    \includegraphics[width=0.11\textwidth]{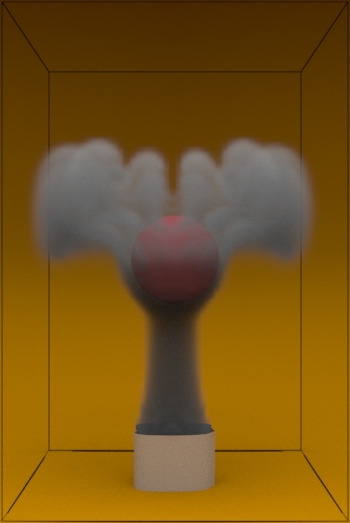}
    \vspace{1pt}
    \\
    \hspace{-4pt}
    \stackunder {\includegraphics[width=0.11\textwidth]{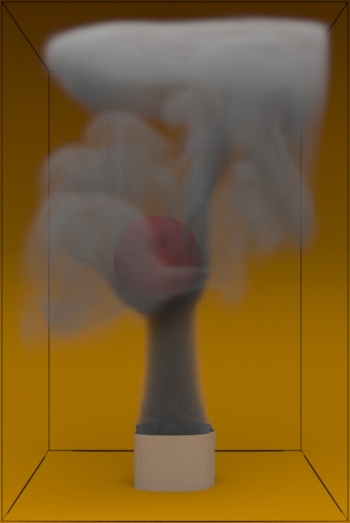}} {\scriptsize{CNN $p_x = 0.44$}}
    \stackunder {\includegraphics[width=0.11\textwidth]{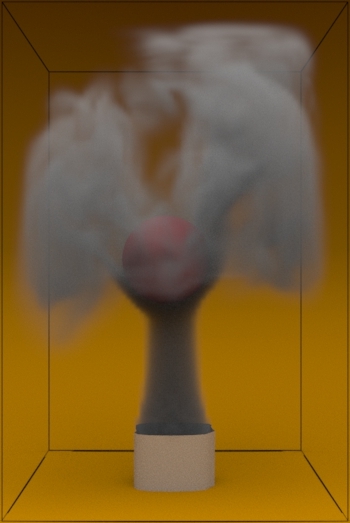}} {\scriptsize{CNN $\hat{p}_x = 0.47$}}
    \stackunder {\includegraphics[width=0.11\textwidth]{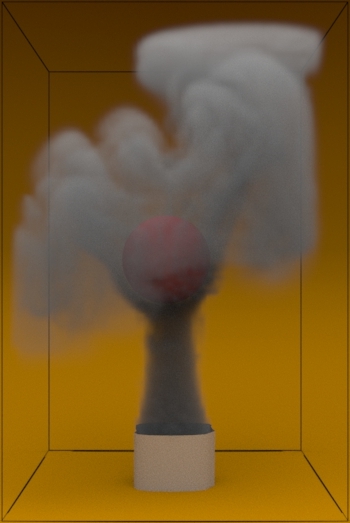}} {\scriptsize{G.t. $p_x = 0.47$}}
    \stackunder {\includegraphics[width=0.11\textwidth]{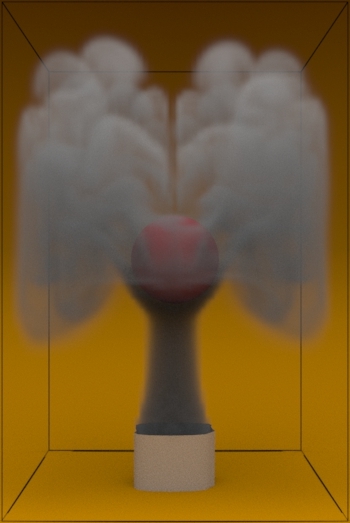}} {\scriptsize{CNN $p_x = 0.5$}}
    \caption{
    Interpolated result (second column) given two input simulations (left and right) with different obstacle positions on the x-axis.
   Our method results in plausible in-betweens compared to ground truth (third column) even for large differences in the input.
   %
    }
    \label{fig:smokeObstacle3D}
    \vspace{-10pt}
\end{figure}

\paragraph{Smoke \& Sphere Obstacle} \Fig{smokeObstacle3D} shows a \threeD~example of a smoke plume interacting with a sphere computed on a grid of size $64\times96\times64$.
The training data consists of ten simulations with varying sphere positions, with the spaces between spheres centroid samples consisting of 0.06 in the interval $[0.2, 0.8]$.
The left and right columns of \Fig{smokeObstacle3D} show the CNN-reconstructed simulations at positions $\mathbf{p}_x = 0.44$ and $\mathbf{p}_x = 0.5$, while the second column presents the interpolated results using our generative network at $\hat{\mathbf{p}}_x = 0.47$. Even with a sparse and hence challenging training data set, flow structures are plausibly reconstructed and compare favorably with ground truth simulations (third column) that were not present in the original \dataset.

\paragraph{Smoke Inflow and Buoyancy} A collection of simulations with varying inflow speed (columns) and buoyancy (rows) is shown in \Fig{smokeGunArray} for the ground truth (left) and our generative network (right). We generated 5 inflow velocities (in the range $[1.0,5.0]$) along with 3 different buoyancy values (from $6\times10^{-4}$ to $1\times10^{-3}$) for 250 frames. Thus, the network was trained with $3,750$ unique velocity fields. \Fig{smokeVaryingSpeedBuoyancy} demonstrates an interpolation example for the buoyancy parameter. The generated simulations on the left and right (using a buoyancy of $6\times10^{-4}$ and $1\times10^{-3}$) closely correspond to the original \dataset~samples, while the second simulation is reconstructed by our CNN using an interpolated buoyancy of $8\times10^{-4}$. We show the ground truth simulation on the third image for a reference comparison.
Our method recovers structures accurately, and the plume shape matches the reference ground truth.

\begin{figure}[h]
    \centering
    \stackunder{\includegraphics[width=0.11\textwidth]{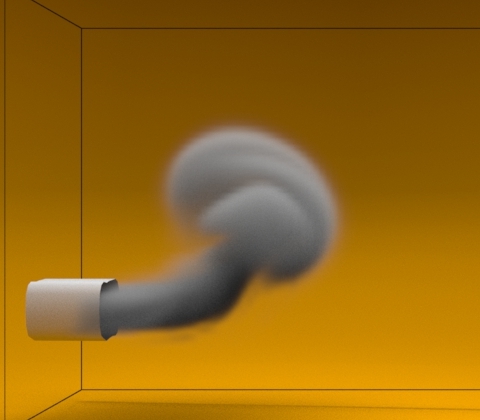}}{\scriptsize{CNN $b = 6 \times 10^{-4}$}}
    \stackunder{\includegraphics[width=0.11\textwidth]{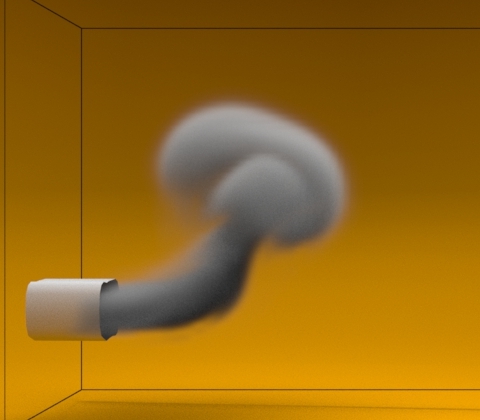}}{\scriptsize{CNN $\hat{b} = 8 \times 10^{-4}$}}
    \stackunder{\includegraphics[width=0.11\textwidth]{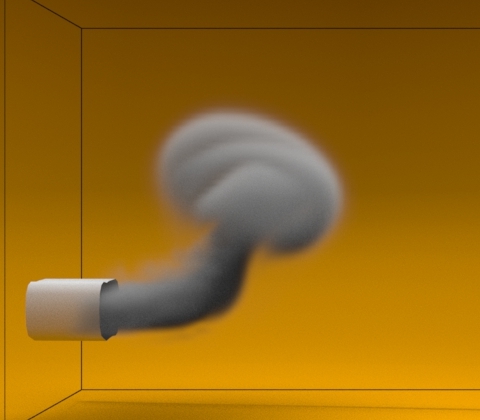}}{\scriptsize{G.t. ${b} = 8 \times 10^{-4}$}}
    \stackunder{\includegraphics[width=0.11\textwidth]{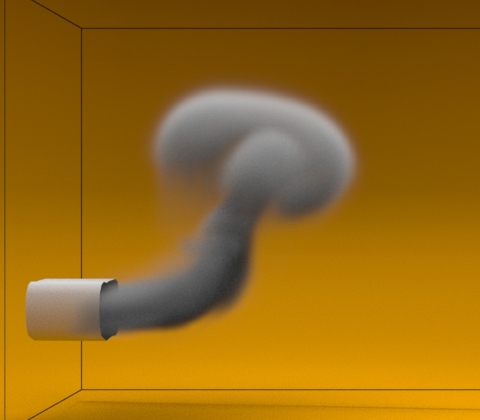}}{\scriptsize{CNN $b = 1 \times 10^{-3}$}}
      \caption{
      Reconstructions of the rising plume scene (left and right), reconstruction for an interpolated buoyancy value ($\hat{b} = 8 \times 10^{-4}$) (second image) and the corresponding ground truth (third image).
    }
    \label{fig:smokeVaryingSpeedBuoyancy}
    \label{fig:smokeVaryingSpeedBuoyancy}
    \vspace{-12pt}
\end{figure}

\begin{figure}[h]
    \centering 
    \stackunder {\includegraphics[trim={0 20 0 10},clip,width=0.11\textwidth]{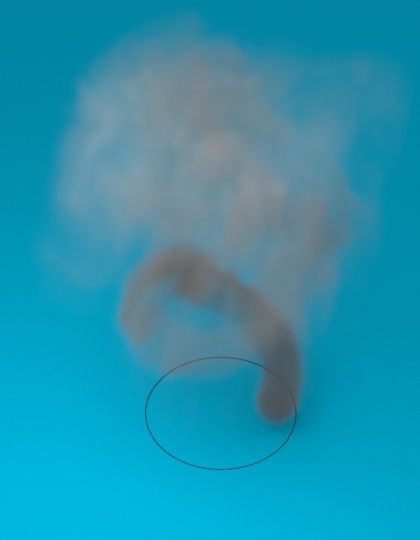}} {\scriptsize{G.t., last frame}}
    \stackunder {\includegraphics[trim={0 20 0 10},clip,width=0.11\textwidth]{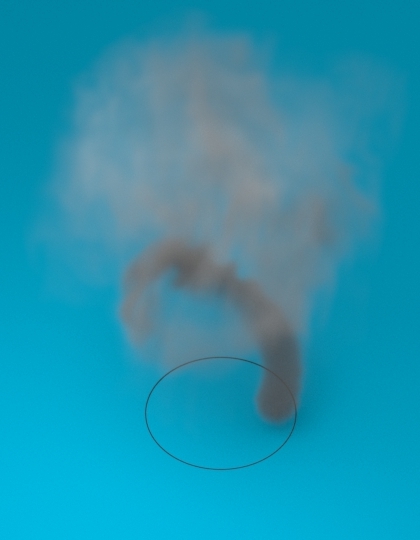}} {\scriptsize{$+20\%$}}
    \stackunder {\includegraphics[trim={0 20 0 10},clip,width=0.11\textwidth]{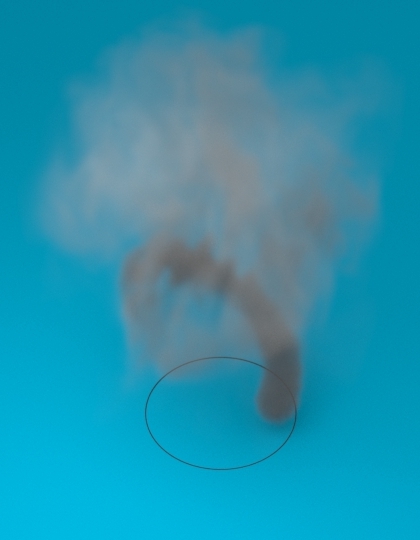}} {\scriptsize{$+60\%$}}
    \stackunder {\includegraphics[trim={0 20 0 10},clip,width=0.11\textwidth]{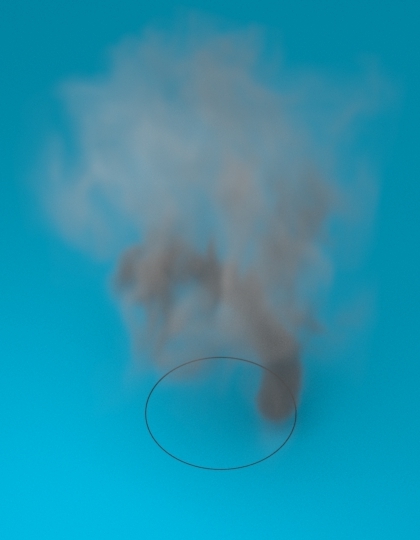}} {\scriptsize{$+100\%$}}
    \caption{
   Time extrapolation results using our latent space integration network. The left image shows the last frame of the ground truth simulation. The subsequent images show results with time extrapolation of $+20\%$, $+60\%$ and $+100\%$ of the original frames.}
    \label{fig:rotsim}
\end{figure}

\paragraph{Rotating Smoke}
We trained our autoencoder and latent space integration network for a smoke simulation with a periodically rotating source using 500 frames as training data. The source rotates in the $XZ$-plane with a period of 100 frames. This example is designed as a stress test for extrapolating time using our latent space integration network. In~\Fig{rotsim}, we show that our approach is able to correctly capture the periodicity present in the original \dataset. Moreover, the method successfully generates another 500 frames, resulting in a simulation that is $100\%$ longer than the original data.

\begin{figure}[t!]
    \centering
    \includegraphics[trim={0 20 0 10},clip,width=0.46\textwidth]{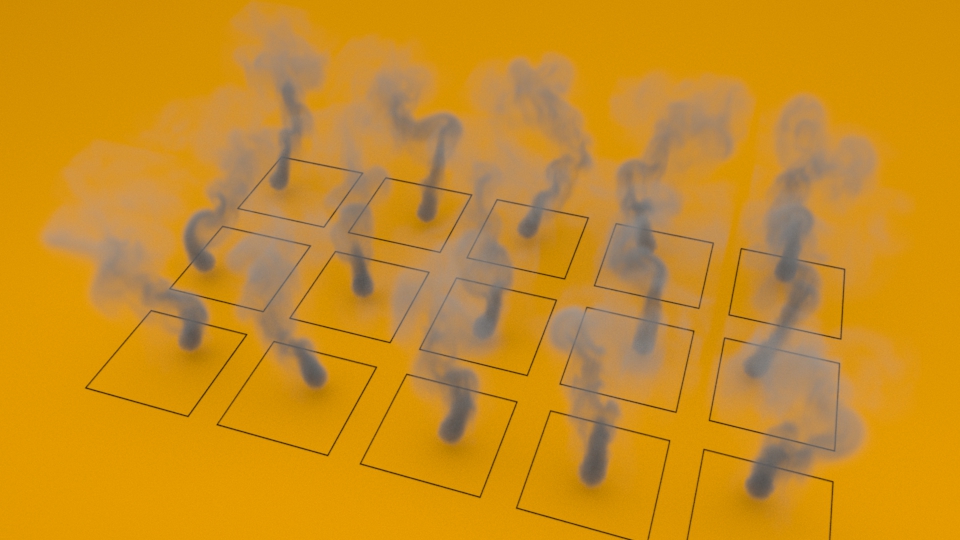}
    \caption{Example simulations of the moving smoke scene used for training the extended parameterization networks.}
    \label{fig:torchgt}
\end{figure}

\begin{figure}[t!]
    \centering 
    \stackunder {\includegraphics[width=0.09\textwidth]{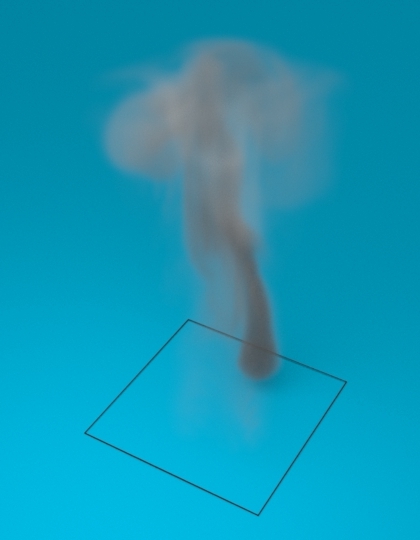}} {\scriptsize{$t=380$}}
    \stackunder {\includegraphics[width=0.09\textwidth]{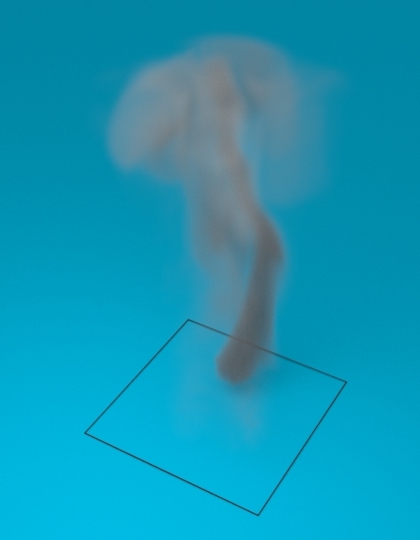}} {\scriptsize{$t=400$}}
    \stackunder {\includegraphics[width=0.09\textwidth]{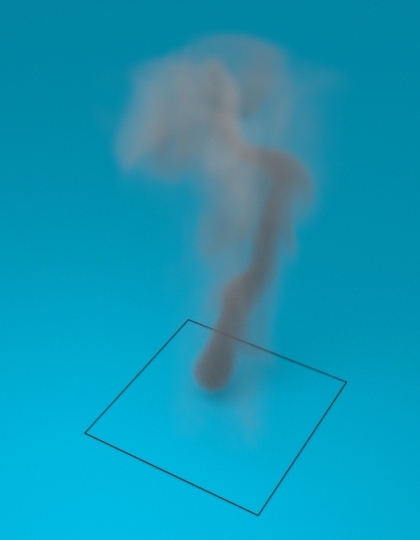}} {\scriptsize{$t=420$}}
    \stackunder {\includegraphics[width=0.09\textwidth]{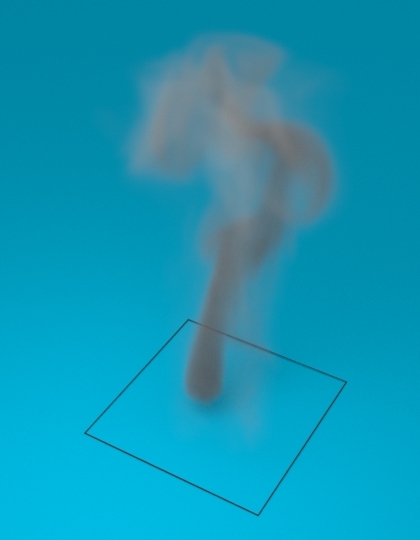}} {\scriptsize{$t=440$}}
    \stackunder {\includegraphics[width=0.09\textwidth]{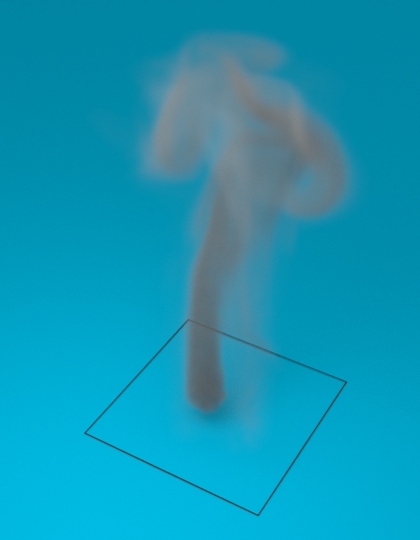}} {\scriptsize{$t=460$}}
    \caption{Different snapshots of a moving smoke source example simulated in the latent space. 
    }
    \label{fig:torchresim}
    \vspace{-15pt}
\end{figure}

\paragraph{Moving Smoke}
A smoke source is moved in the $XZ$-plane along a path randomly generated using Perlin noise. We sampled 200 simulations on a grid of size $48 \times 72 \times 48$ for 400 frames - a subset is shown in \Fig{torchgt} - and used them to train our autoencoder and latent space integration networks.
In \Fig{torchresim}, we show a moving smoke source whose motion is not part of the training data and was computed by integrating in the latent space. We extrapolate in time to increase the simulation duration by $100\%$ (\ie 800 frames). The network generates a plausible flow for this unseen motion over the full course of the inferred simulation. Although the results shown here were rendered offline, the high performance of our trained model would allow for interactive simulations. 

\subsection{\threeD~Liquid Examples}
\begin{figure}[t]
    \centering
    \stackunder{\includegraphics[width=0.11\textwidth]{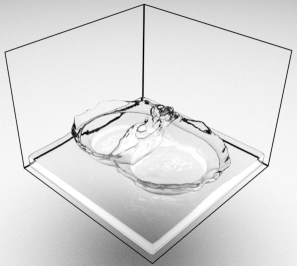}}{\scriptsize{$d = 0.15, \theta=0^{\circ}$}}
    \stackunder{\includegraphics[width=0.11\textwidth]{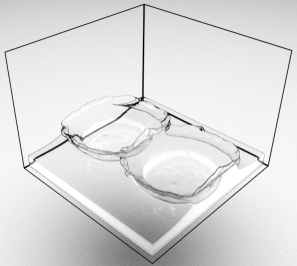}}{\scriptsize{$d = 0.25, \theta=0^{\circ}$}}
    \stackunder{\includegraphics[width=0.11\textwidth]{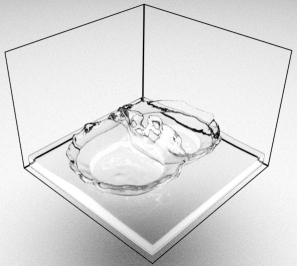}}{\scriptsize{$d = 0.15, \theta=90^{\circ}$}}
    \stackunder{\includegraphics[width=0.11\textwidth]{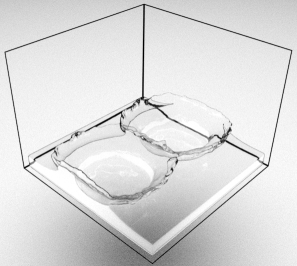}}{\scriptsize{$d = 0.25, \theta=90^{\circ}$}}
    \caption{Training samples for the liquid spheres scene. In total we used $50$ simulation examples with varying distances and angles.}
    \label{fig:3dLiquidDrop}
\end{figure}

\begin{figure}[t]
    \centering
    \stackunder{\includegraphics[width=0.14\textwidth]{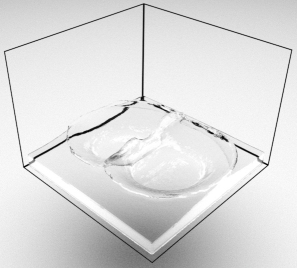}}{\scriptsize{$d = 0.15, \theta=0^{\circ}$}}
    \stackunder{\includegraphics[width=0.14\textwidth]{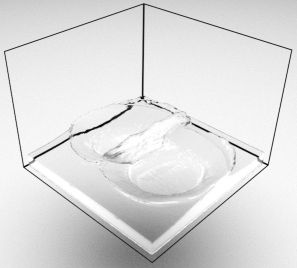}}{\scriptsize{$\hat{d} = 0.1625, \hat{\theta}=9^{\circ}$}}
    \stackunder{\includegraphics[width=0.14\textwidth]{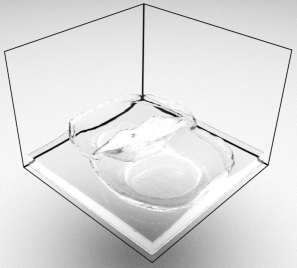}}{\scriptsize{$d = 0.175, \theta=18^{\circ}$}}
    \caption{
   CNN-generated results with parameter interpolation for the liquid spheres example. While the far left and right simulations employ parameter settings that were part of the training data, the middle example represents a new in-between parameter point which is successfully reconstructed by our method.
	}
    \label{fig:3dLiquidInterp}
    \vspace{-15pt}
\end{figure}

\paragraph{Spheres Dropping on a Basin} We demonstrate that our approach can also handle splashing liquids.
We use a setup for two spheres dropping on a basin, which is parameterized by the initial distance of the spheres, as well as by the initial drop angles along the $XZ-$plane relative to the basin. We sample velocity field sequences by combining 5 different distances and 10 different angles; \Fig{3dLiquidDrop} shows $4$ of the training samples.
With $150$ frames in time, the network is trained with $7,500$ unique velocity fields. We used a single-phase solver and extrapolated velocities from the liquid to the air phase before training (extrapolation size $= 4$ cells).
\Fig{3dLiquidInterp}, middle, shows our result for an interpolated angle of $\hat{\theta} = 9^\circ$ and a sphere distance of $\hat{d} = 0.1625$, given two CNN-reconstructed input samples on the left ($\theta = 0^\circ, d = 0.15$) and right ($\theta = 18^\circ, d = 0.175$).
Our results demonstrate that the bulk of the liquid dynamics are preserved well. Small scale details such as high-frequency structures and splashes,
however, are particularly challenging and deviate from the reference simulations.

\begin{figure}[h]
    \centering
    \includegraphics[width=0.22\textwidth]{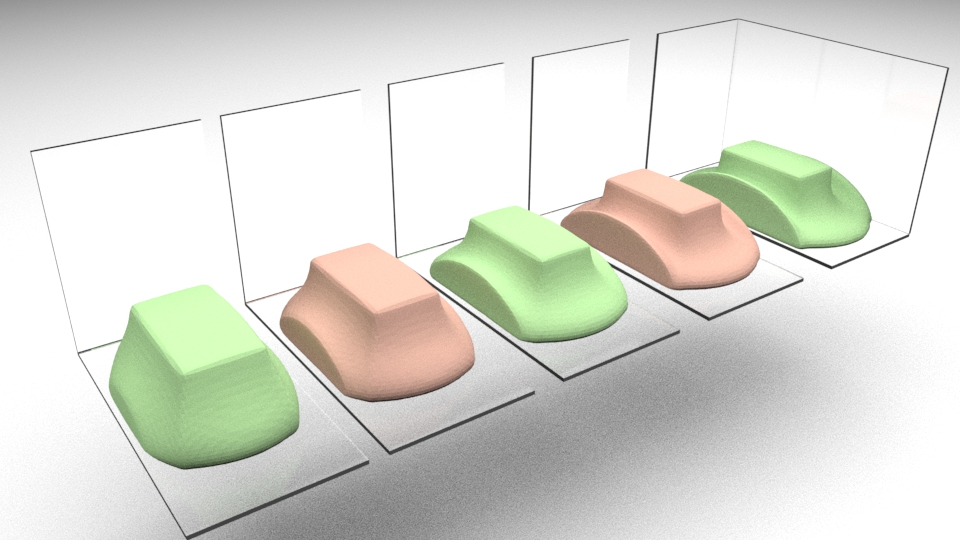}
    \includegraphics[width=0.22\textwidth]{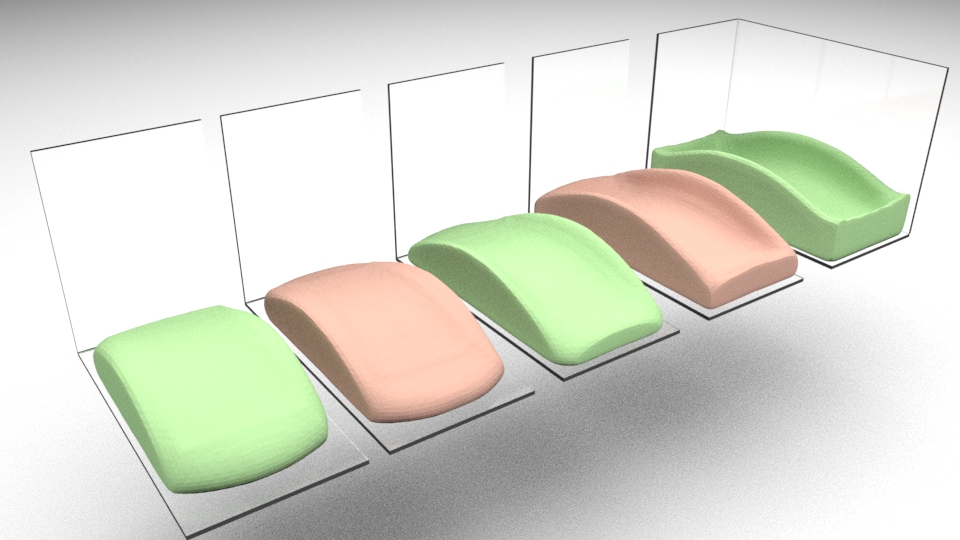}
    \caption{
    Snapshots of a CNN reconstructed dam break with different viscosity strengths for two different frames. Green liquids denote correspondences with ground truth ($\mu = 2\times[10^{-4}, 10^{-3}, 10^{-2}]$, back to front) while pink ones are interpolated ($\hat{\mu} = 2\times[5^{-3}, 5^{-2}]$, back to front).}
    \label{fig:liquidViscosity}
\end{figure}

\begin{figure}[h]
    \centering
    \stackunder{\includegraphics[width=0.20\textwidth]{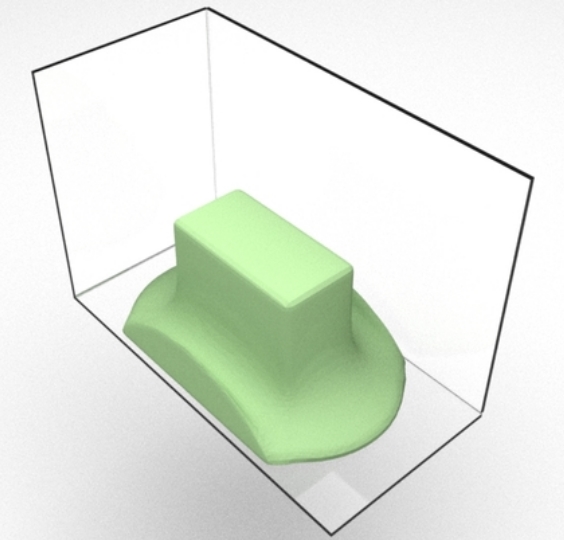}}{\scriptsize{Frame 23, $\hat{\mu}=1.04 \times 10^{-3}$}}
    \stackunder{\includegraphics[width=0.20\textwidth]{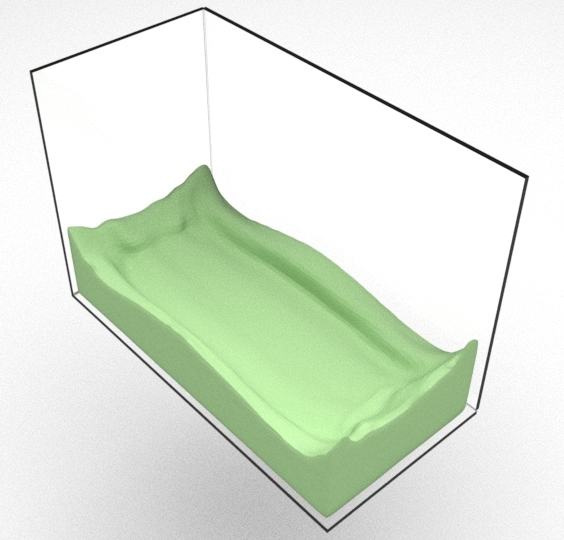}}{\scriptsize{Frame 64, $\hat{\mu}=7.14 \times 10^{-3}$}}
    \caption{Reconstruction result using a time varying viscosity strength. In the first few frames the liquid quickly breaks into the container. As the simulation advances, the viscosity increases and the liquid sticks to a deformed configuration.}
    \label{fig:liquidVaryingViscosity}
    \vspace{-15pt}
\end{figure}

\paragraph{Viscous Dam Break}
In this example, a dam break with four different viscosity strengths $(\mu = 2\times[10^{-5}, 10^{-4}, 10^{-3}, 10^{-2}])$ was used to train the network. Our method can reconstruct simulations with different viscosities accurately,
and also interpolate between different viscosities with high fidelity. In \Fig{liquidViscosity}, the CNN-reconstructed, green-colored liquids have direct correspondences in the original dataset; the pink-colored simulations are interpolation results between the two nearest green samples. Additionally, it is also possible to increase the viscosity over time as shown in \Fig{liquidVaryingViscosity}.
The results show that this works reliably although the original parameterization does neither support time-varying viscosities nor do the training samples represent such behavior.

\paragraph{Slow Motion Fluids}
Our supplemental video additionally shows an interesting use case that is enabled by our CNN-based interpolation: the generation of temporally upsampled simulations.
Based on a trained model we can create slow-motion effects, which we show for the liquid drop and dam break examples.

\section{Evaluation and Discussion}
\label{sec:Evaluation}
\subsection{Training}\label{sec:Training}
Our networks are trained on normalized data in the range $[-1, 1]$. In case of velocity fields, we normalize them by the maximum absolute value of the entire data set. We found that batch or instance normalization techniques do not improve our velocity fields output, as the highest (lowest) pixel intensity and mean deviation might vary strongly within a single batch. Frames from image-based data sets have a uniform standard deviation, while velocity field snapshots can vary substantially. Other rescaling techniques, such as standardization or histogram equalization, could potentially further improve the training process. 

\paragraph{Convergence of the Training}
The presented architecture is very stable and all our tests have converged reliably. Training time highly depends on the example and the targeted reconstruction quality. Generally, \threeD~liquid examples require more training iterations (up to 100 hours of training) in order to get high quality surfaces, while our smoke examples finished on average after 72 hours of training.

\begin{figure}[h]
    \centering
    \includegraphics[width=0.475\textwidth]{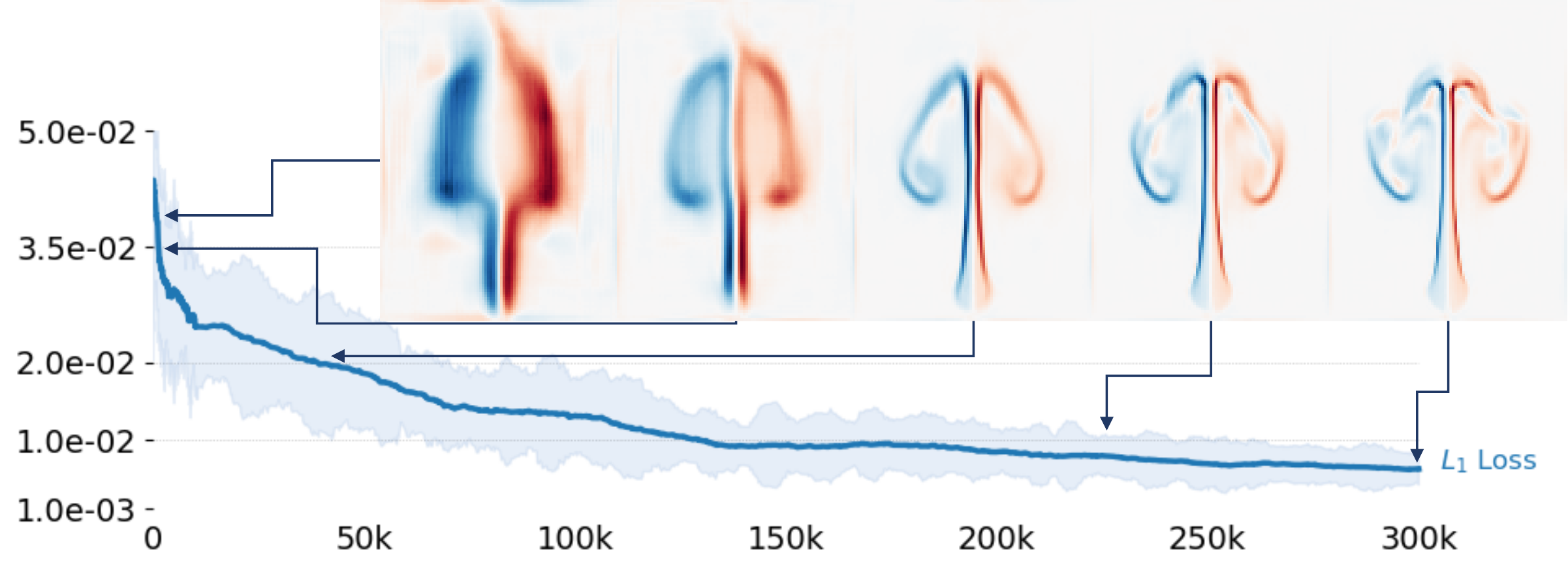}
    \caption{Convergence plot of the $L_1$ loss for the \twoD~smoke sequence from \Fig{smokeVorticityInterpolation}.}
    \label{fig:progress}
    \vspace{-15pt}
\end{figure}

\Fig{progress} shows a convergence plot of the \twoD~smoke example, with training iterations on the $x$-axis and error on the $y$-axis. The superimposed images show clearly how quality increases along with training iterations. After about $180,000$ iterations, the smoke plume is already reconstructed with good accuracy. This corresponds to roughly $3$ hours of training on our hardware. Scaling tests with various \twoD~grid resolutions ($64\times48$, $128\times96$, $256\times192$, $512\times384$) have shown that the training speed scales proportionally with the resolution, while keeping the mean absolute error at a constant level.

\subsection{Performance Analysis}

Table~\ref{tab:stat} summarizes the statistics of all presented examples. In terms of wall-clock time, the proposed CNN approach generates velocity fields up to \maxSpeedup~faster than re-simulating the data with the underlying CPU solver. Some care must be taken when interpreting this number because our Tensorflow network runs on the GPU, while the original Mantaflow code runs on the CPU. Fluid simulations are known to be memory bandwidth-limited \cite{Kim:2008:HAO}, and the bandwidth discrepancy between a GTX 1080 (320 GB/s) and our Intel desktop (25.6 GB/s) is a factor of 12.5. However, even if we conservatively normalize by this factor, our method achieves a speed-up of up to $58\times$. Thus, the core algorithm is still at least an order of magnitude faster. To facilitate comparisons with existing subspace methods, we do not include the training time of our CNN when computing the maximum speedup, as precomputation times are customarily reported separately. Instead, we include them in the discussion of training times below.


Contrary to traditional solvers, our approach is able to generate multiple frames in time independently. Thus, we can efficiently concatenate CNN queries into a GPU batch, which then outputs multiple velocity fields at once. Adding more queries increases the batch size (Table~\ref{tab:stat}, 5th column, number in brackets), and the maximum batch size depends on the network size and the hardware's memory capacity. Since we are using the maximum batch size possible for each scene, the network evaluation time scales inversely with the maximum batch size supported by the GPU. Due to the inherent capability of GPUs to efficiently schedule floating point operations, the time for evaluating a batch is independent of its size or the size of the network architecture. 
Additionally, our method is completely oblivious to the complexity of the solvers used to generate the data. Thus, more expensive stream function \cite{Ando2015} or energy-preserving \cite{Mullen2009} solvers could potentially be used with our approach, yielding even larger speed-ups.

In contrast, computing the linear basis using traditional SVD-based subspace approaches can take between 20 \cite{Kim2013} and 33 \cite{Wicke2009} hours. The process is non-iterative, so interrupting the computation can yield a drastically inferior result, i.e.~the most important singular vector may not have been discovered yet. \citet{Stanton:2013:NGP} reduced the precomputation time to 12 hours, but only by using a 110-node cluster. In contrast, our iterative training approach is fully interruptible and runs on a single machine.

\paragraph{Compression}
The memory consumption of our method is at most 30 MB, which effectively compresses the input data by up to \maxCompression. Previous subspace methods \cite{Jones2016} only achieved ratios of $14\times$, hence our results improve on the state-of-the-art by two orders of magnitude.
We have compared the compression ability of our network to FPZIP~\cite{lindstrom2006fast}, a data reduction technique often used in scientific visualization~\cite{meyer2017data}. FPZIP is a prediction-based compressor using Lorenzo predictor, which is designed for large floating-point \datasets~in arbitrary dimension. For the two data sets of \Sec{smokePlume2D} and \Sec{3dsmoke}, we reached a compression of $172\times$ and $356\times$, respectively. In comparison, FPZIP achieves a compression of $4\times$ for both scenes with 16 bits of precision (with a mean absolute error comparable to our method). 
When allowing for a $6\times$ larger mean absolute error, FPZIP achieves a $47\times$ and $39\times$ compression. I.e., 
the data is more than $4\times$ larger and has a reduced quality compared to our encoding. Thus, our method outperforms commonly used techniques for compressing scientific data. Details can be found in the supplemental material.

\begin{table*}
        \begin{tabular}{p{2cm} cc c c c r r r r}
                \hline
                 & \small{Grid} &   & \small{Simulation} & \small{Eval. Time} & \textbf{\small{Speed Up}} & \small{Data Set} & \small{Network} & \textbf{\small{Compression}} & \small{Training}\\
                \small{Scene} & \small{Resolution} & \# \small{Frames} & \small{Time (s)} & \small{(ms) [Batch]} & \small{($\times$)} & \small{Size (MB)} & \small{Size (MB}) & \textbf{\small{Ratio}} & \small{Time (h)}				
		\tabularnewline \hline
		\small Smoke Plume & $96\times128$ &  21,000 & 0.033 & 0.052 [100] & 635 & 2064 & 12 & 172 & 5 \\
		\small Smoke Obstacle & $64 \times 96 \times 64$ & 6,600 & 0.491 & 0.999 [5] & 513 & 31143 & 30 & 1038 & 74\\
		\small Smoke Inflow & $112 \times 64 \times 32$   & 3,750 & 0.128 & 0.958 [5] & 128 & 10322 & 29 & 356 & 40\\
		\small Liquid Drops & $96 \times 48 \times 96$ & 7,500 & 0.172 & 1.372 [3] & 125 & 39813 & 30 & \textbf{1327} & 134\\
		\small Viscous Dam & $96 \times 72 \times 48$ & 600 & 0.984 & 1.374 [3] & \textbf{716} & 2389 & 29 & 82 & 100\\
		\small Rotating Smoke & $48 \times 72 \times 48$ & 500 & 0.08 & 0.52 [10] & 308 & 995 & 38 & 26 & 49\\
		\small Moving Smoke & $48 \times 72 \times 48$ & 80,000 & 0.08 & 0.52 [10] & 308 & 159252 & 38 & $4191^*$ & 49
		\tabularnewline \hline
	\end{tabular}
	\caption{Statistics for training \datasets~and our CNN. Note that simulation excludes advection and is done on the CPU, while network evaluation is executed on the GPU with batch sizes noted in brackets. In case of liquids, the conjugate gradient residual threshold is set to $1e^{-3}$, while for smoke it is $1e^{-4}$. For the Rotating and Moving Smoke scenes, the numbers for training time and network size include both the autoencoder and latent space integration networks. \newline * We optimize the network for subspace simulations rather than the quality of reconstruction, so we do not take this number into account when evaluating the maximal compression ratio.}
	\label{tab:stat}
\vspace{-12pt}
\end{table*}

\subsection{Quality of Reconstruction and Interpolation}
\label{sec:qualityReconstruction}
\setlength{\columnsep}{0pt}%
\begin{wrapfigure}[7]{hR}{0.4\linewidth} 
\vspace{-3mm}
\begin{center}
\includegraphics[trim=15px 0px 10px 7px, clip, width=0.4\linewidth]{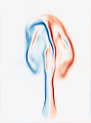}
\includegraphics[trim=15px 0px 10px 7px, clip, width=0.4\linewidth]{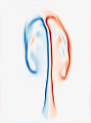}
\end{center}
\label{fig:insetGradComparision}
\end{wrapfigure}
\paragraph{Training Data}
Several factors affect the reconstruction and interpolation quality. An inherent problem of machine learning approaches is that quality strongly depends on the data used for training. In our case, the performance of our generative model for interpolated positions is sensitive to the input sampling density and parameters. If the sampling density is too coarse, or if the output abruptly changes with respect to the variation of parameters, errors may appear on reconstructed velocity fields. These errors include the inability to accurately reconstruct detailed flow structures, artifacts near obstacles, and especially ghosting effects in the interpolated results. An example of ghosting is shown in the inset image where only 11 training samples are used (left), instead of the 21 (right) from \Sec{smokePlume2D}.

\setlength{\columnsep}{0pt}%
\begin{wrapfigure}[6]{hR}{0.45\linewidth} 
\vspace{-3.5mm}
\begin{center}
\includegraphics[width=0.43\linewidth]{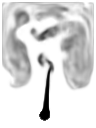}
\includegraphics[width=0.43\linewidth]{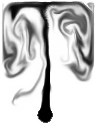}
\end{center}
\end{wrapfigure}

\paragraph{Target Quantities}
We have also experimented with training directly with density values (inset image, left) instead of the velocity fields (inset image, right). In case of density-trained networks, the dynamics fail to recover the non-linear nature of momentum conservation and artifacts appear. Advecting density with the reconstructed velocity field yields significantly better results. A more detailed discussion about the quality of the interpolation regarding the number of input samples and discrepancies between velocity and density training is presented in the supplemental material.

\begin{figure}[h]
    \centering
	\stackunder{\includegraphics[width=0.078\textwidth]{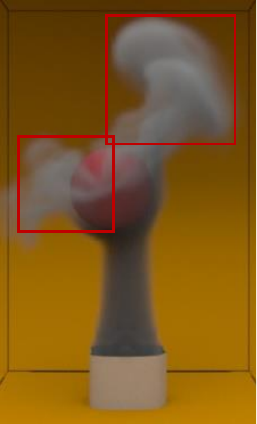}} {\scriptsize{$\hat{\vec{u}}_\vec{c} = G(\vec{c})$}}
	\stackunder{\includegraphics[width=0.078\textwidth]{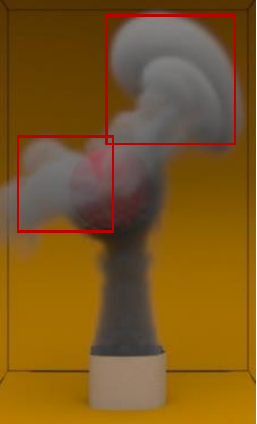}} {\scriptsize{$\hat{\vec{u}}_\vec{c} = \nabla \times G(\vec{c})$}}
	\stackunder{\includegraphics[width=0.078\textwidth]{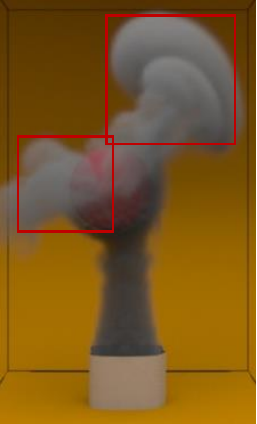}} {\scriptsize{G.t.}}
	\stackunder{\includegraphics[width=0.198\textwidth]{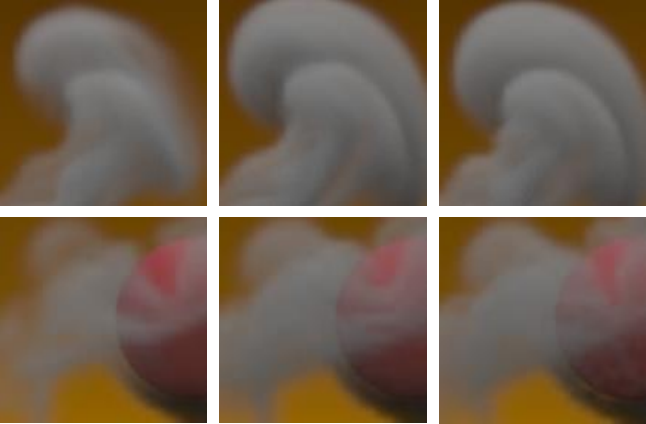}} {\scriptsize{Close-up views}}
    \caption{Comparisons of the results from networks trained on our compressible loss, incompressible loss and the ground truth, respectively. On the right sequence we show the highlighted images from the simulations on the left. We notice that the smoke patterns from the incompressible loss are closer to ground truth simulations.}
    \label{fig:comp_incomp}
    \vspace{-8pt}
\end{figure}

\paragraph{Velocity Loss}
A comparison between our compressible loss, incompressible functions and ground truth is shown in \Fig{comp_incomp}. The smoke plume trained with the incompressible loss from \Eq{lossdivfree} shows a richer density profile closer to the ground truth, compared to results obtained using the compressible loss.

\begin{figure}[h]
    \centering 
    \includegraphics[trim={0 50 0 0}, clip, width=0.085\textwidth]{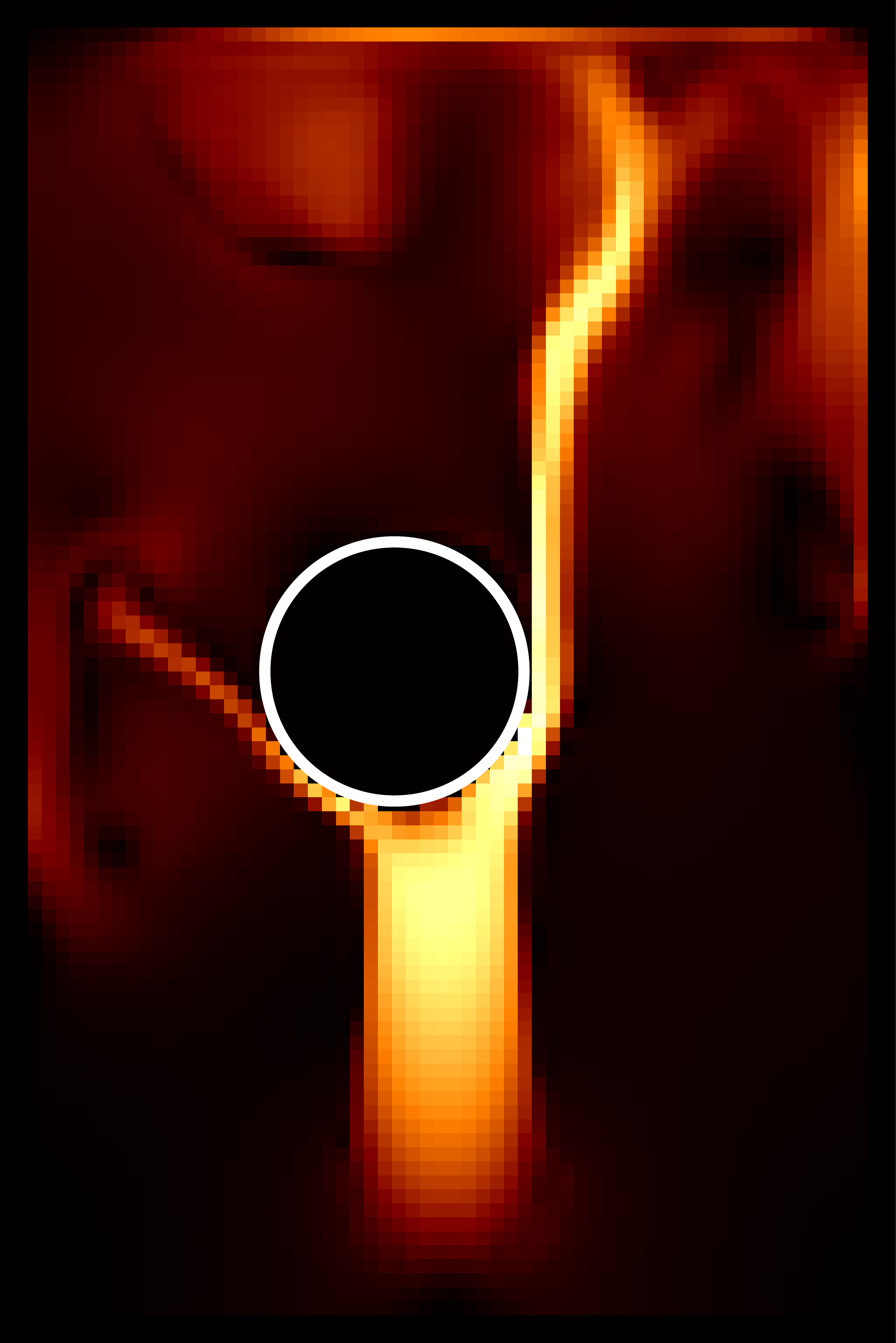}
    \includegraphics[trim={0 50 0 0}, clip, width=0.085\textwidth]{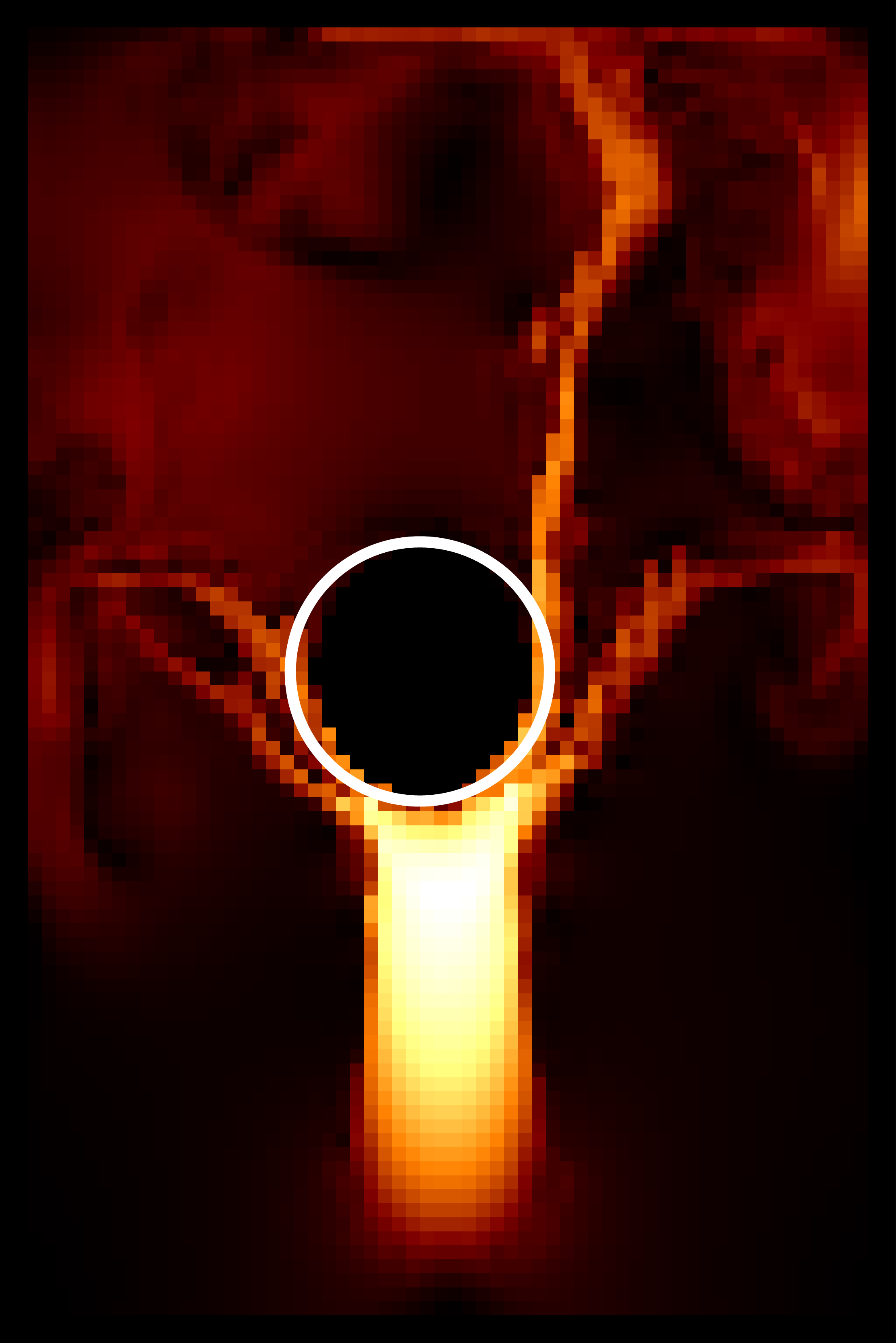}
    \includegraphics[trim={0 50 0 0}, clip, width=0.085\textwidth]{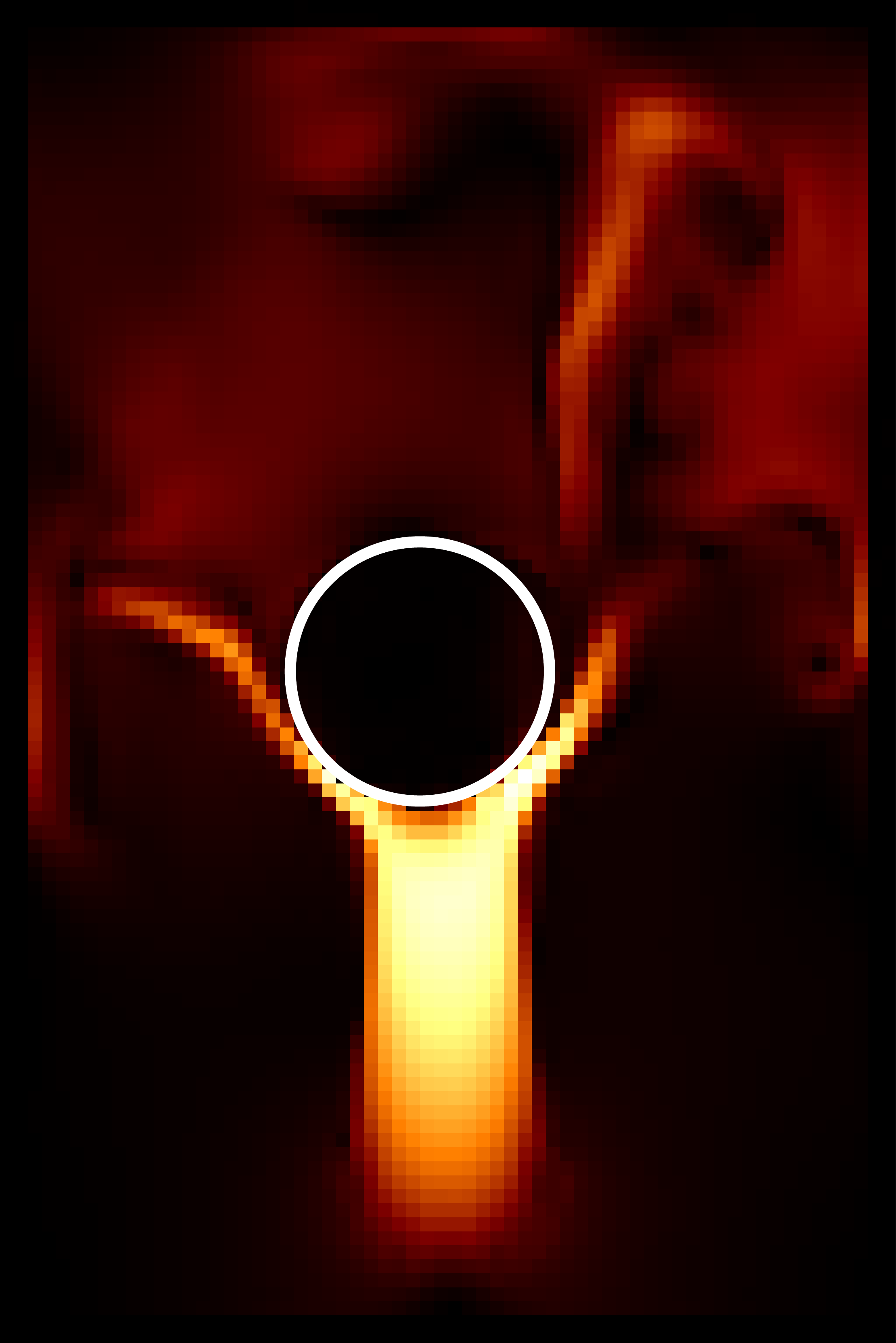}
    \includegraphics[trim={0 50 0 0}, clip, width=0.085\textwidth]{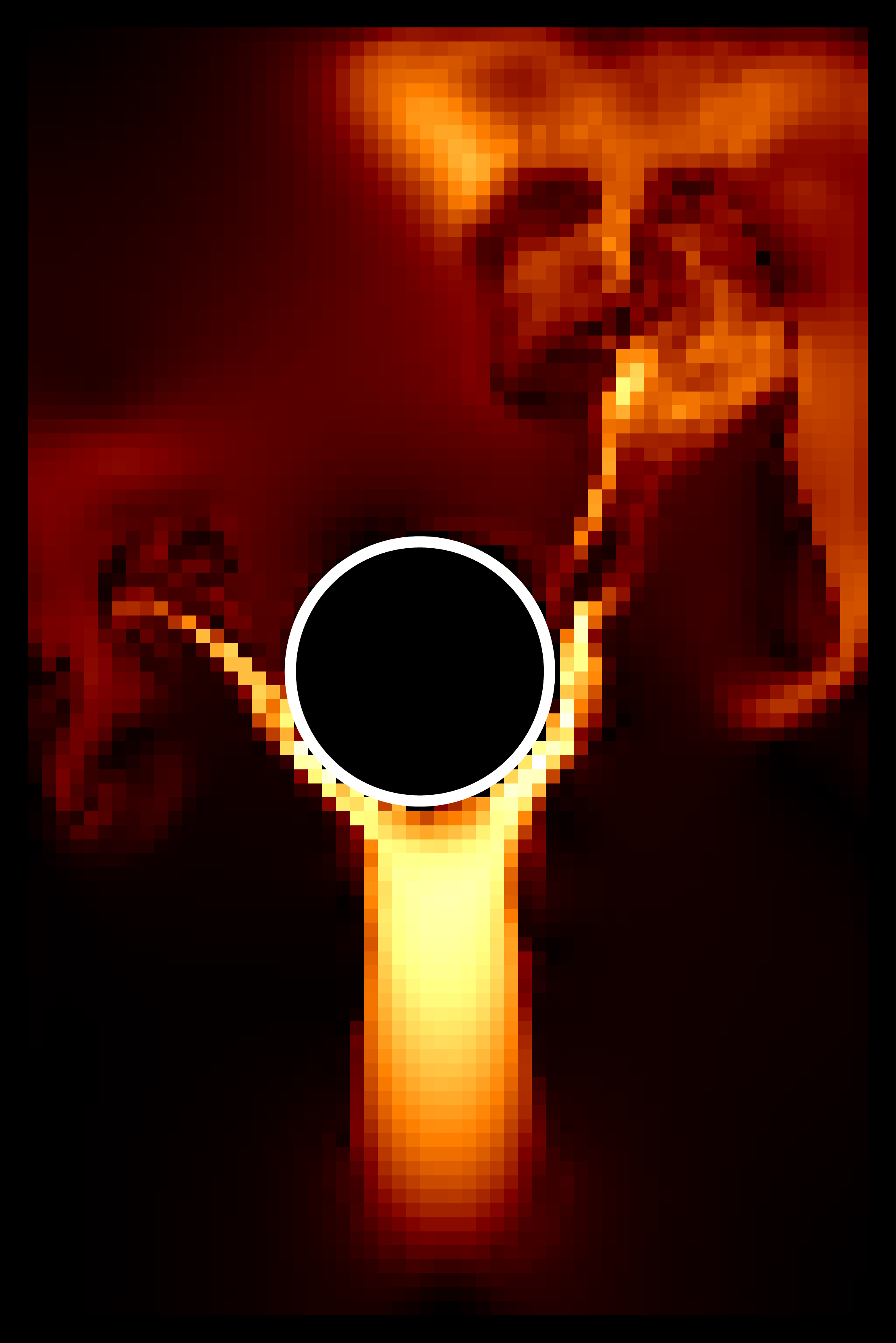}
    \includegraphics[trim={0 50 0 0}, clip, width=0.085\textwidth]{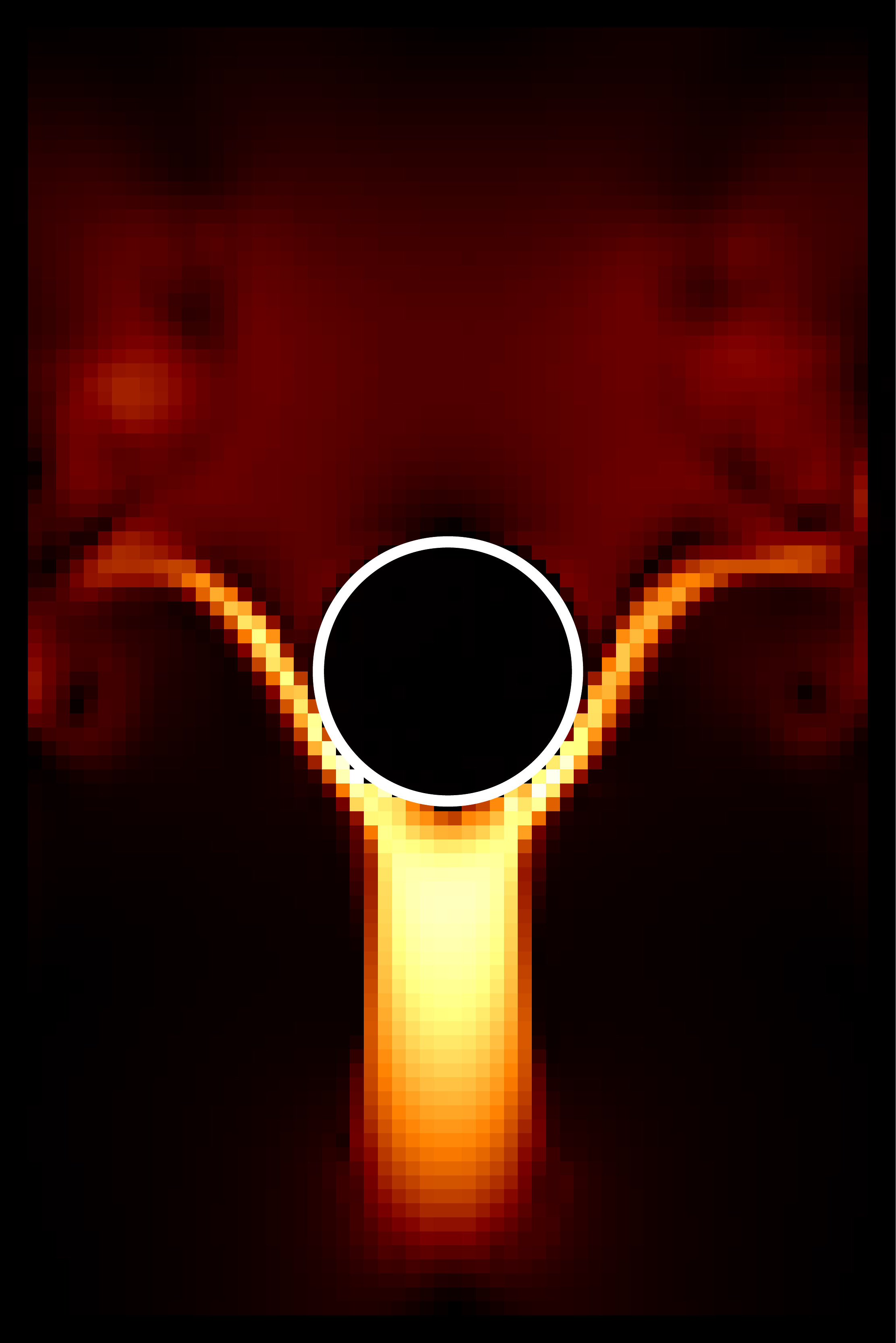}
    \vspace{1pt}
    \\
    \hspace{-4pt}
    \stackunder {\includegraphics[trim={0 80 0 0}, clip, width=0.085\textwidth]{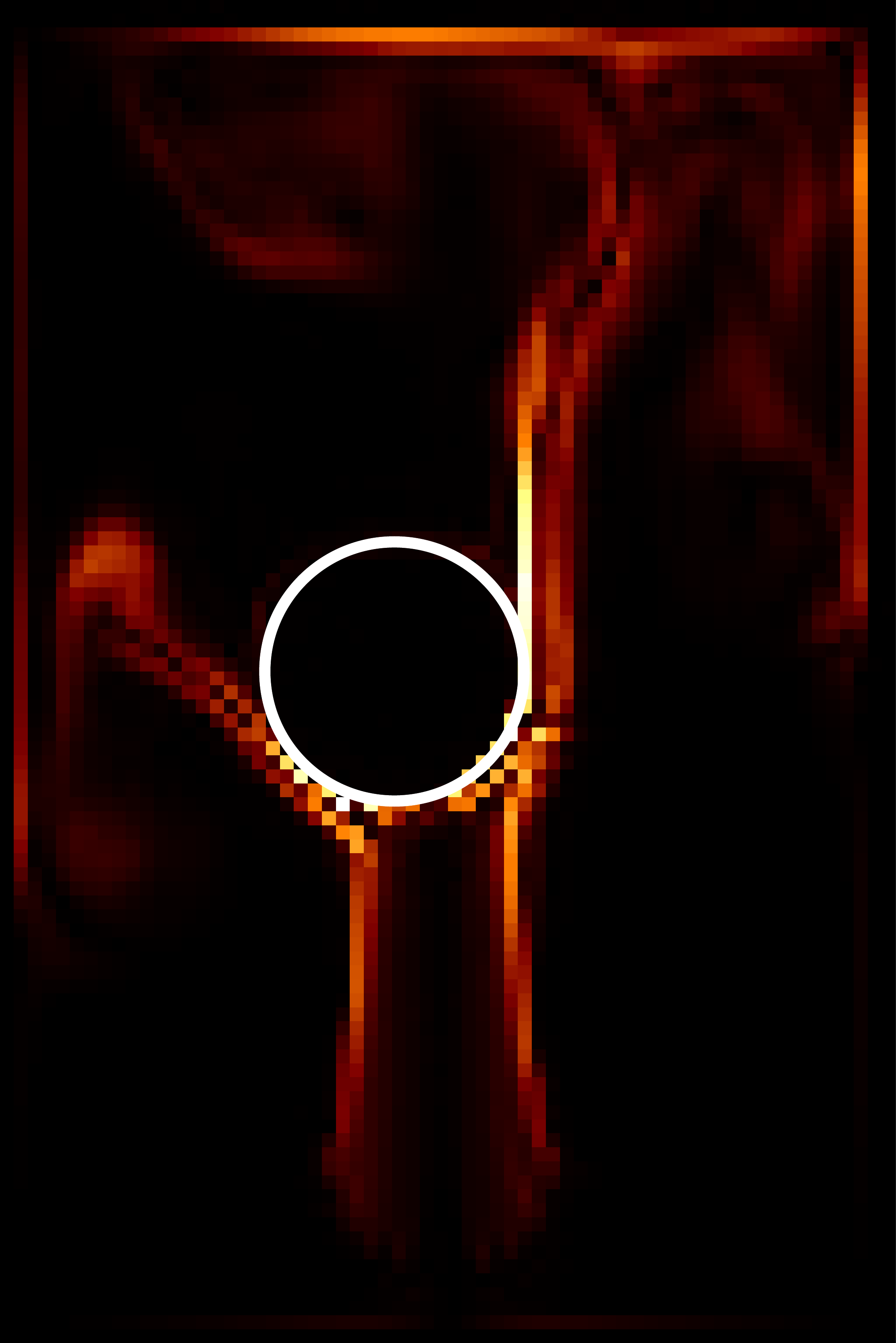}} {\scriptsize{CNN $p_x = 0.44$}}
    \stackunder {\includegraphics[trim={0 80 0 0}, clip, width=0.085\textwidth]{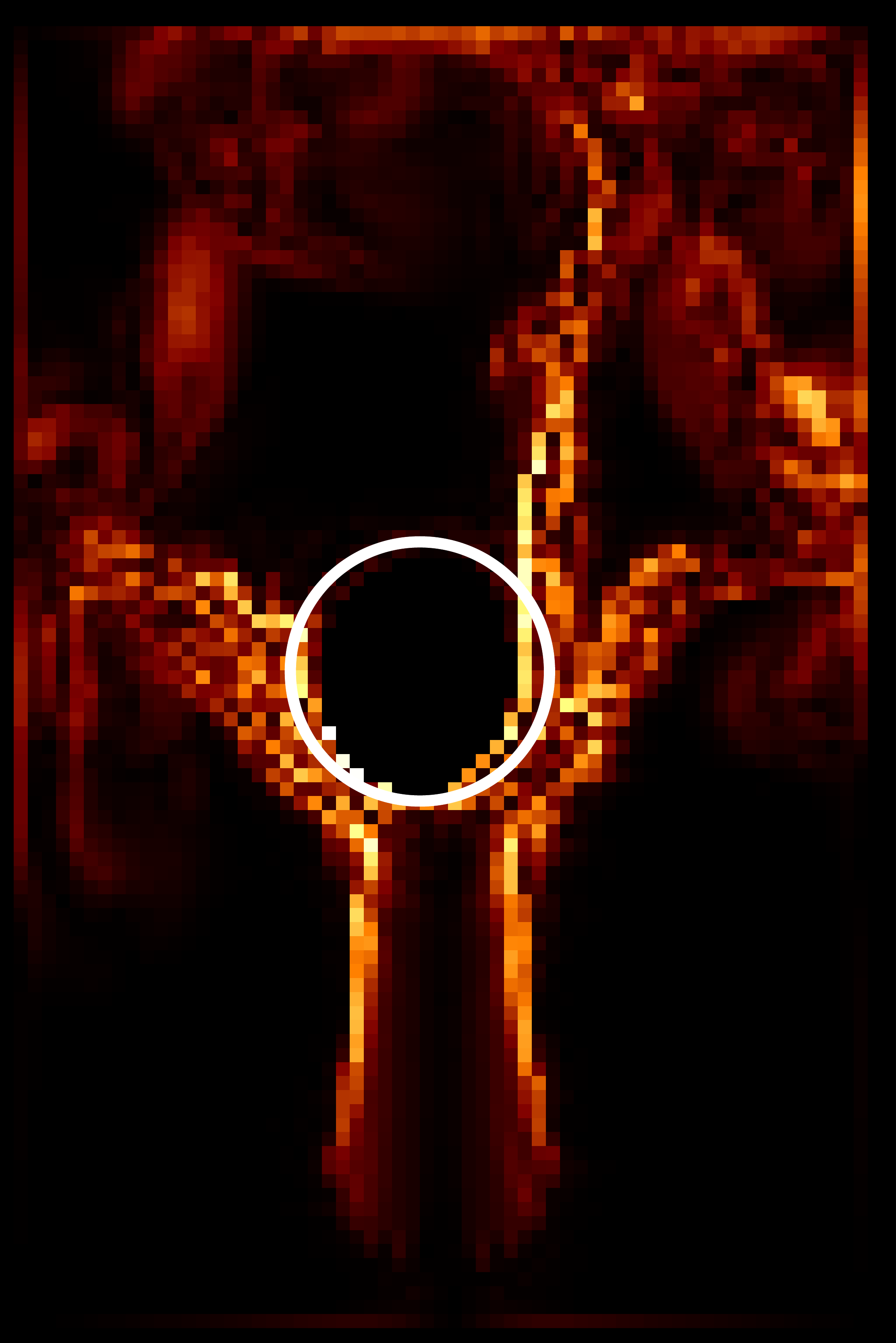}} {\scriptsize{Lin. $\hat{p}_x = 0.47$}}
    \stackunder {\includegraphics[trim={0 80 0 0}, clip, width=0.085\textwidth]{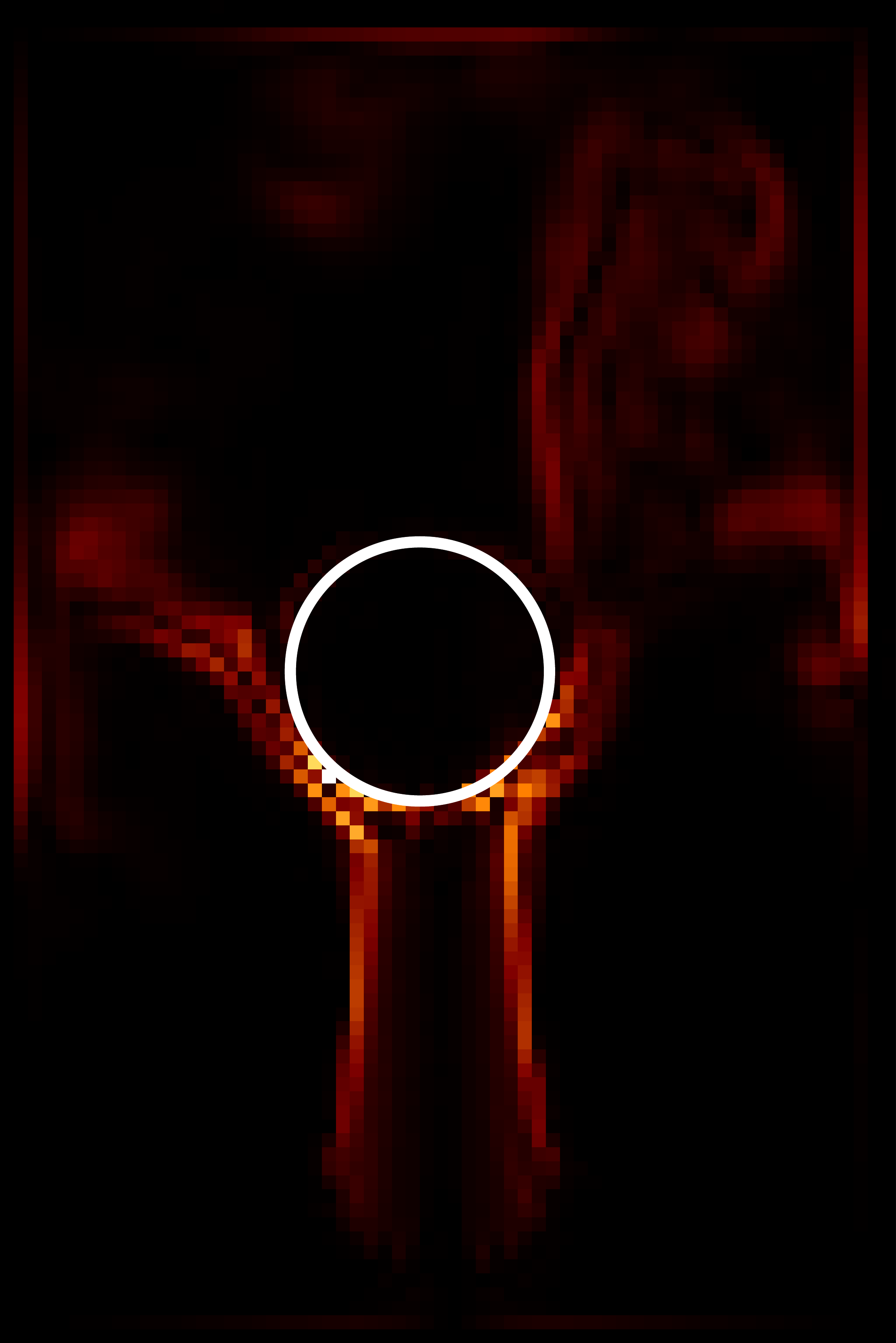}} {\scriptsize{CNN $\hat{p}_x = 0.47$}}
    \stackunder {\includegraphics[trim={0 80 0 0}, clip, width=0.085\textwidth]{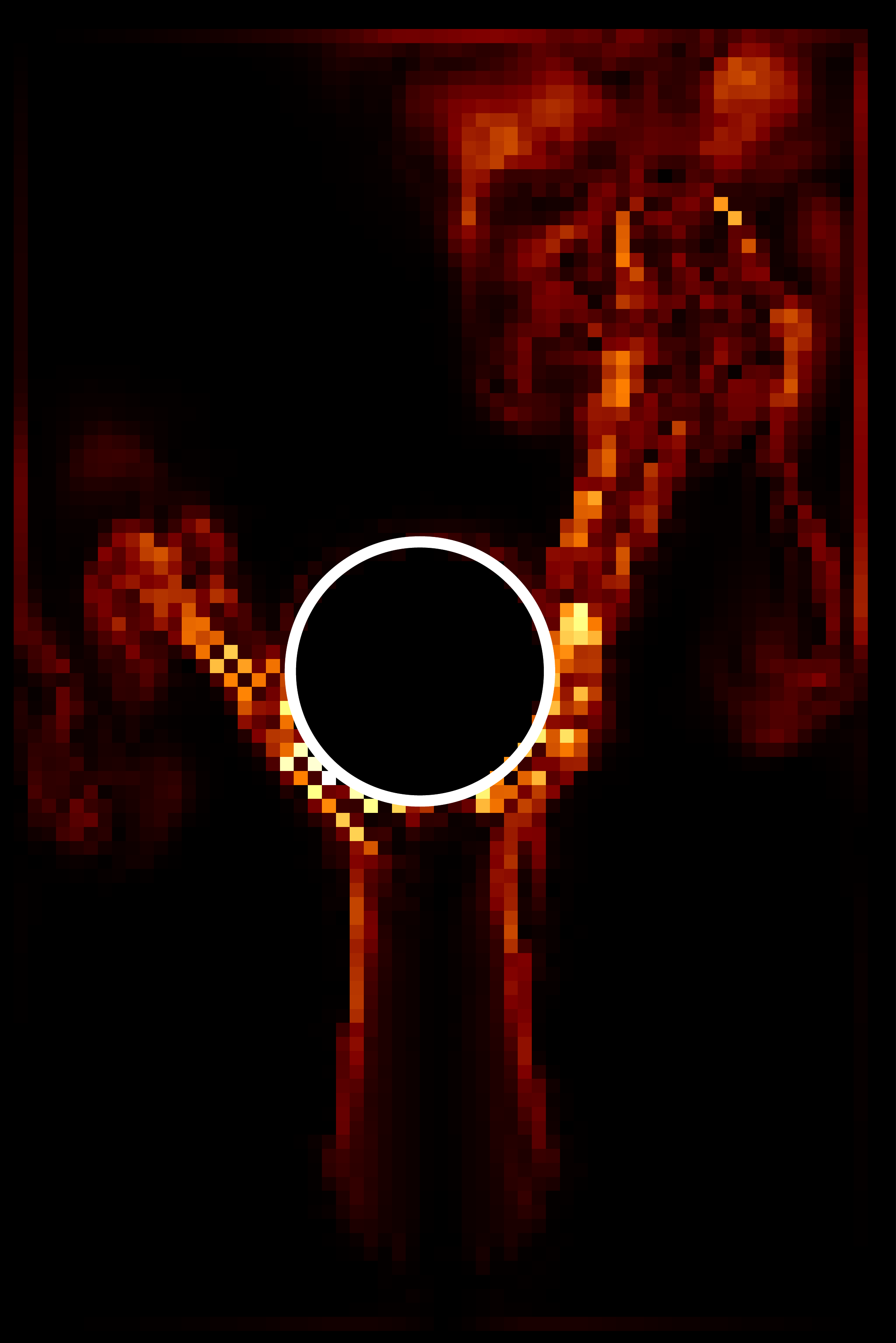}} {\scriptsize{G.t. $p_x = 0.47$}}
    \stackunder {\includegraphics[trim={0 80 0 0}, clip, width=0.085\textwidth]{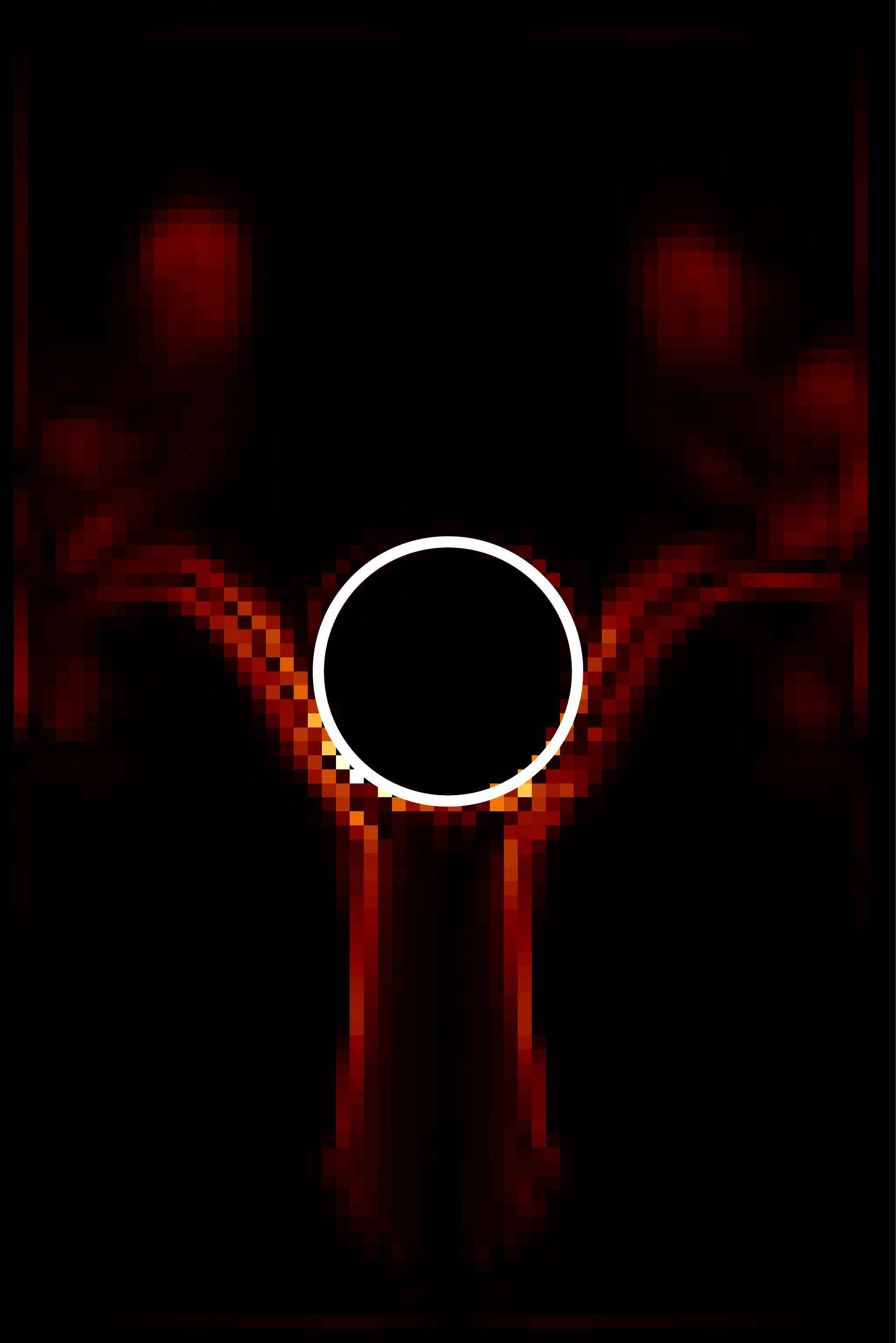}} {\scriptsize{CNN $p_x = 0.5$}}
    \caption{Slice views of the last row of Fig.~\ref{fig:smokeObstacle3D}. The color code represents the velocity (top) and vorticity (bottom) magnitudes. The second column shows a linear interpolation of the input. Despite the absence of any constraints on boundary conditions, our method (third column) preserves the shape of the original sphere obstacle,
    and yields significantly better results than the linear interpolation.}
    \label{fig:smokeObstacle3Dslice}
\end{figure}

\begin{figure}[h]
    \centering
    \includegraphics[trim={0 10 0 8}, clip,width=0.44\textwidth]{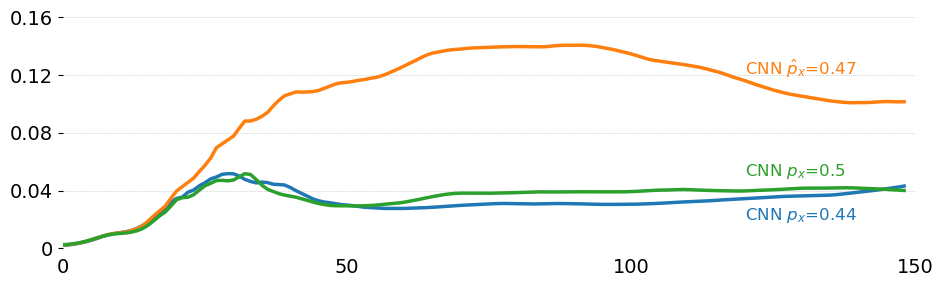}
    \caption{Mean absolute error plot of velocity penetration for the smoke obstacle example.
    %
    Though errors in interpolated samples are a bit higher than those of reconstructed samples,
    they do not exceed 1\% of the maximum absolute value 
   of the \dataset.}
    \label{fig:penetration}
    \vspace{-15pt}
\end{figure}

\paragraph{Boundary Conditions} The proposed CNN is able to handle immersed obstacles and boundary conditions without additional modifications.
\Fig{smokeObstacle3Dslice} shows sliced outputs for the scene from \Fig{smokeObstacle3D} which contains a sphere obstacle.
We compare velocity (top) and vorticity magnitudes (bottom).
The first and last images show the reconstruction of the CNN for $p_\vec{x}$ positions that have correspondences in the training \dataset. The three images in the middle show results from linearly blending the closest velocity fields, our CNN reconstruction and the ground truth simulation, from left to right respectively. In the case of linearly blended velocity fields, ghosting arises as
features from the closest velocity fields are super-imposed~\cite{Thuerey2016}, and the non-penetration constraints for the obstacle are not respected, as velocities are present inside the intended obstacle positions. In \Fig{penetration}, we plot the resulting velocity penetration errors. Here we compute the mean absolute values of the velocities inside the voxelized sphere, normalized by the mean sum of the velocity magnitudes for all cells around a narrow band of the sphere.
Boundary errors are slightly higher for interpolated parameter regions (orange line in \Fig{penetration}), since no explicit constraint for the object's shape is enforced. However, the regularized mean error still accounts for less than 1\% of the maximum absolute value of the velocity field. Thus, our method successfully preserves the non-penetration boundary conditions.

\paragraph{Liquid-air Interface} Due to the separate advection calculation for particles in our FLIP simulations, smaller splashes can leave the velocity regions generated by our CNNs, causing surfaces advected by reconstructed velocities to hang in mid-air.
Even though the reconstructed velocity fields closely match the ground truth samples, liquid scenes are highly sensitive to such variations. We removed FLIP particles that have smaller velocities than a threshold in such regions, which was sufficient to avoid hanging particles artifacts.

\subsection{Extrapolation and Limitations}

\begin{figure}[t]
    \centering
    \includegraphics[trim={0 10 0 0},clip,width=0.085\textwidth]{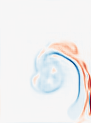}
    \includegraphics[trim={0 10 0 0},clip,width=0.085\textwidth]{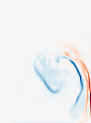}
    \includegraphics[trim={0 10 0 0},clip,width=0.085\textwidth]{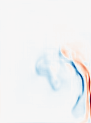}
    \includegraphics[trim={0 10 0 0},clip,width=0.085\textwidth]{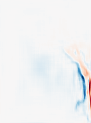}
    \includegraphics[trim={0 10 0 0},clip,width=0.085\textwidth]{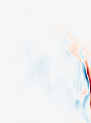}
    \vspace{1pt}
    \\
    \includegraphics[trim={0 5 0 0},clip,width=0.085\textwidth]{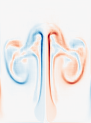}
    \includegraphics[trim={0 5 0 0},clip,width=0.085\textwidth]{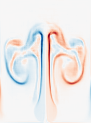}
    \includegraphics[trim={0 5 0 0},clip,width=0.085\textwidth]{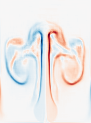}
    \includegraphics[trim={0 5 0 0},clip,width=0.085\textwidth]{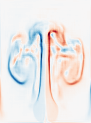}
    \includegraphics[trim={0 5 0 0},clip,width=0.085\textwidth]{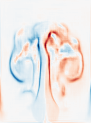}
    \vspace{1pt}
    \\
    \hspace{-4pt}
    \stackunder {\includegraphics[width=0.085\textwidth]{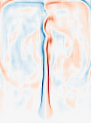}} {\scriptsize{Last}}
    \stackunder {\includegraphics[width=0.085\textwidth]{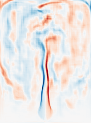}} {\scriptsize{$+5\%$}}
    \stackunder {\includegraphics[width=0.085\textwidth]{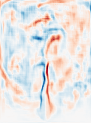}} {\scriptsize{$+10\%$}}
    \stackunder {\includegraphics[width=0.085\textwidth]{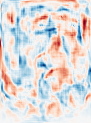}} {\scriptsize{$+20\%$}}
    \stackunder {\includegraphics[width=0.085\textwidth]{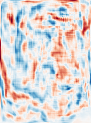}} {\scriptsize{$+30\%$}}
    \caption{\twoD~smoke plume extrapolation results (from t. to b.: position, inflow width, time) where only the generative network is used.
    Plausible results can be observed for up to 10\% extrapolation.
    }
    \label{fig:smoke2DExtp}
    \vspace{-15pt}
\end{figure}

\paragraph{Extrapolation with Generative Model}
We evaluated the extrapolation capabilities for the case where only the generative part of our \cnnName~CNN (Section \ref{sec:Method}) is used. Generally, extrapolation works for sufficiently small increments beyond the original parameter space. \Fig{smoke2DExtp} shows an experiment in which we used weights that were up to 30\% of the original parameter range ($[-1, 1]$). The leftmost images show the vorticity plot for the maximum value of the range for the position (top), inflow size (middle), and time (bottom) parameters of the \twoD~smoke plume example. The rightmost images show the maximum variation of parameters, in which the simulations deteriorate in quality. In practice, we found that up to $10\%$ of extrapolation still yielded plausible results.

\paragraph{Limitations}
Our \cnnName~CNN is designed to generate velocity fields for parameterizable scenes. As such, our method is not suitable for reconstructing arbitrary velocity fields of vastly different profiles by reduction to a shared latent representation. As discussed in \Sec{qualityReconstruction}, there is also no enforcement of physical constraints such as boundary conditions for intermediate interpolated parameters. Thus, the capability of the network to reconstruct physically accurate samples on interpolated locations depends on the proximity of the data samples in the parameter space. Additionally, the reconstruction quality of the autoencoder and latent space integration networks are affected by the size of the latent space $\vec{c}$, and there is a possible issue of temporal jittering because of lack of the gradient loss on~\Eq{losstn}. We provide an extended discussion regarding the reconstruction quality and the latent space size on our supplemental material.

\section{Conclusion}
We have presented the first generative deep learning architecture that successfully synthesizes plausible and divergence-free \twoD~and {\threeD} fluid simulation velocities from a set of reduced parameters. Our results show that generative neural networks are able to construct a wide variety of fluid behaviors, from turbulent smoke to viscous liquids, that closely match the input training data. Moreover, our network can synthesize physically plausible motion when the input parameters are continuously varied to intermediate states that were not present during training. In addition, we can handle complex parameterizations in a reduced latent space, enabling flexible latent space simulations by using a latent space integration network.

Our solver has a constant evaluation time and is considerably faster (up to \maxSpeedup) than 
simulations with the underlying CPU solver, which makes our approach attractive
for re-simulation scenarios where input interactions can be parameterized. These performance characteristics immediately suggest applications in games and virtual environments. As high resolution fluid simulations are also known to demand large disk and memory budgets, the compression characteristics of our algorithm (with up to \maxCompression) render the method attractive for movie productions as well.

Our CNN architecture was carefully designed to achieve high quality fluid simulations, which is why the loss function considers both the velocity field and its gradient. Over the course of evaluating many alternatives, we found that the most important factor to simulation quality was the amount of training data. If the \datasets~are too sparse, artifacts appear, and important flow structures are missing. We address this issue by simply increasing the number of training samples, but in scenarios where data was directly captured or the simulation times are prohibitive, this may not be feasible. Improving the reconstruction quality of interpolated states over sparsely sampled \datasets~is an open direction for future work.

Overall, we found that the proposed CNN is able to reproduce velocity fields accurately. However, for small-scale details or near discontinuities such as boundary conditions,
the network can sometimes smooth out fine flow structures. A possible future research direction is the exploration of generative adversarial networks (GANs), partial convolutions~\cite{Liu2018}, joint training with other fields such as SDF, or alternative distance measures to enhance the accuracy for fine structures in the data. Additionally, it is an interesting direction to improve the latent space integration network in various ways, such as using an end-to-end training with the autoencoder or the gradient loss for temporal smoothness.

Generally, combining fluid simulations and machine learning is largely unexplored, and more work is needed to evaluate different architectures and physics-inspired components.
Our CNN-based algorithm is the first of its kind in the fluids community, and we believe that the speed, interpolation and compression capabilities can enable a variety of future applications, such as interactive liquid simulations \cite{Prantl2017} or implementation of fluid databases.

\section{Acknowledgments}
\label{sec:Acknowledgments}
This work was supported by the Swiss National Science Foundation (Grant No. 200021\_168997) and ERC Starting Grant 637014.


\bibliographystyle{eg-alpha-doi}

\bibliography{deepfluid}

\end{document}